\documentclass[lettersize,journal]{IEEEtran}
\usepackage{algorithmic}
\usepackage{algorithm}
\usepackage{hyperref}
\usepackage{multirow}
\usepackage{amsmath}
\usepackage{tikz}
\usepackage{graphicx} 
\usepackage{tikz}
\usepackage{enumitem}
\usepackage{amsmath}
\usepackage{cite}
\usepackage{multirow}
\usepackage{algorithm}
\usepackage{algorithmic}
\usepackage{enumitem}
\usepackage{amsmath,amssymb,amsfonts}
\usepackage{graphicx}
\usepackage[english]{babel}
\usepackage{amsthm}
\usepackage{amssymb}
\usepackage{amssymb}
\usepackage{array}
\usepackage[caption=false,font=normalsize,labelfont=sf,textfont=sf]{subfig}
\usepackage{textcomp}
\usepackage{stfloats}
\usepackage{url}
\usepackage{dsfont}
\usepackage{verbatim}
\usepackage{xcolor}
\usepackage{graphicx}
\usepackage{soul}
\usepackage{cite}
\usepackage{amsmath,amssymb,amsfonts}
\usepackage{theoremref}

\newtheorem{thm}{Theorem}

\DeclareMathOperator{\dist}{dist}
\usepackage{amsmath,amsfonts}
\usepackage{algorithmic}
\usepackage{array}
\usepackage[caption=false,font=normalsize,labelfont=sf,textfont=sf]{subfig}
\usepackage{textcomp}
\usepackage{stfloats}
\usepackage{url}
\usepackage{verbatim}
\usepackage{graphicx}
\def\BibTeX{{\rm B\kern-.05em{\sc i\kern-.025em b}\kern-.08em
    T\kern-.1667em\lower.7ex\hbox{E}\kern-.125emX}}
\usepackage{balance}
\begin{document}
\title{Semi-decentralized Inference in Heterogeneous Graph Neural Networks for Traffic Demand Forecasting: An Edge-Computing Approach}


\author{Mahmoud Nazzal, ~\IEEEmembership{Member,~IEEE,}, Abdallah Khreishah,~\IEEEmembership{Senior
Member,~IEEE,}, Joyoung Lee, Shaahin Angizi,~\IEEEmembership{Senior Member,~IEEE,}, Ala Al-Fuqaha,~\IEEEmembership{Senior
Member,~IEEE}, and Mohsen Guizani,~\IEEEmembership{Fellow,~IEEE,}\\

\thanks{
M. Nazzal, A. Khreishah, and S. Angizi are with the Department of Electrical and Computer Engineering, New Jersey Institute of Technology, Newark,
NJ 07102 USA (e-mail: \{mn69,abadallah,shaahin.angizi\}@njit.edu)
\\
J. Lee, is with the the Department of Civil Engineering, New Jersey Institute of Technology, Newark,
NJ 07102 USA (e-mail: jo02@njit.edu)
\\
Ala Al-Fuqaha is with the Hamad Bin Khalifa University, Doha 34110, Qatar
(e-mail: aalfuqaha@hbku.edu.qa).
\\
M. Guizani is with the Machine Learning Department, Mohamed Bin
Zayed University of Artificial Intelligence (MBZUAI), Abu Dhabi, UAE (email: mguizani@ieee.org)
}
}

\markboth{Journal of \LaTeX\ Class Files,~Vol.~18, No.~9, September~2020}%
{How to Use the IEEEtran \LaTeX \ Templates}

\maketitle

\begin{abstract}
Prediction of taxi service demand and supply is essential for improving customer's experience and provider’s profit. Recently, graph neural networks (GNNs) have been shown promising for this application. This approach models city regions as nodes in a transportation graph and their relations as edges. GNNs utilize local node features and the graph structure in the prediction. However, more efficient forecasting can still be achieved by following two main routes; enlarging the scale of the transportation graph, and simultaneously exploiting different types of nodes and edges in the graphs. However, both approaches are challenged by the scalability of GNNs. An immediate remedy to the scalability challenge is to decentralize the GNN operation. However, this creates excessive node-to-node communication. In this paper, we first characterize the excessive communication needs for the decentralized GNN approach. Then, we propose a semi-decentralized approach utilizing multiple cloudlets, moderately sized storage and computation devices, that can be integrated with the cellular base stations. This approach minimizes inter-cloudlet communication thereby alleviating the communication overhead of the decentralized approach while promoting scalability due to cloudlet-level decentralization. Also, we propose a heterogeneous GNN-LSTM algorithm for improved taxi-level demand and supply forecasting for handling dynamic taxi graphs where nodes are taxis. Extensive experiments over real data show the advantage of the semi-decentralized approach as tested over our heterogeneous GNN-LSTM algorithm. Also, the proposed semi-decentralized GNN approach is shown to reduce the overall inference time by about an order of magnitude compared to centralized and decentralized inference schemes.
\end{abstract}

\begin{IEEEkeywords}
GNN, hetGNN, taxi demand forecasting, taxi supply forecasting, ITS, decentralized inference.
\end{IEEEkeywords}

\section{Introduction}
\label{Section1}

\IEEEPARstart{I}{ntelligent} transportation system (ITS) is an essential item in modern city planning. A key component of an ITS is the means of public transportation such as taxis, buses, and ride-hailing vehicles. As the importance of these services grows, there is a corresponding need for accurately and efficiently forecasting the travel needs of passengers and the corresponding available supplies by vacant taxis ready to serve them. This forecasting enables efficient management of transportation resources and allows for dynamic allocation of taxis such that customer waiting time is minimized and the taxi occupancy times are maximized. It can also help optimizing routes, urban development, traffic flow, and public transportation planning. 

\par Taxi demand forecasting has been receiving increasing amounts of attention in the recent transportation engineering literature \cite{li2012prediction,yu20193d,moreira2013predicting,taxi2017tlc,chen2020multi,xu2019incorporating,liu2019contextualized,davis2020grids,xu2022application}. Similar to other prediction problems, approaches to taxi demand and supply forecasting can be broadly categorized into two main categories; first is model-based approaches where a statistical model for traffic patterns is employed. Examples along this line include integrated auto-regressive moving average (ARIMA) \cite{moreira2013predicting} and linear regression \cite{tong2017simpler} models. Despite their simplicity, these methods focus only on temporal dependencies and overlook exploiting spatial dependencies for the prediction. The other category is deep learning (DL)-based methods where data-driven techniques are shown to well exploit spatiotemporal correlations for improved prediction. Along the DL line, recurrent neural networks (RNN) such as long short-term memory (LSTM) are especially important for taxi demand and supply forecasting as they can well address time dependency. Accordingly, there is a series of works on using RNN for taxi demand and supply forecasting \cite{xu2017real,wang2018deepstcl,ke2017short,liu2019contextualized}.

\par Recently, there has been a growing interest in the use of graph neural networks (GNNs) for taxi demand and supply prediction. This approach models city regions as nodes in a graph and their relations as the edges linking these nodes. Along this line, several works have shown the advantage of GNNs in utilizing local region information and the relationships across non-Euclidean regions in improving the forecasting performance \cite{li2017diffusion,yu20193d,xu2019incorporating,wu2019graph,davis2020grids,chen2020multi,ye2021coupled,xu2022application}. Despite the promising success of GNNs for taxi demand and supply forecasting, there are still outstanding challenges hindering their potential. First, as for graph representation, it is advantageous to simultaneously expose several node and relation types in the representation learning of transportation graphs \cite{hong2020heteta}. Specifically, the existence of multiple types of nodes and edges in current transportation graphs calls for adopting a heterogeneous information network (HIN) approach to seamlessly exploit them. This requires the development of corresponding heterogeneous GNNs (hetGNNs) to handle their representation learning. Second, inferring taxi demand and supply predictions at the level of the whole transportation system incurs a huge amount of computation as it is necessary to utilize the existing salient relationships. This calls for developing solutions to improve the scalability of the GNN approach to cope with city-wide or even larger graphs \cite{lee2021decentralized}.

\par Based on the above discussion, in this paper, we propose a hetGNN-LSTM-based algorithm for taxi demand and supply prediction. Compared to the existing GNN-based approaches, our algorithm defines taxis as nodes in a graph. This allows taxi graphs to be dynamic as opposed to existing approaches assuming nodes as fixed geographical regions. On the other hand, this allows for predicting the demands and supplies for each taxi. Operating this algorithm in a centralized way is computationally intensive. Therefore, we develop a decentralized GNN inference approach. However, we show theoretically and through experiments that this decentralized approach has a huge amount of message passing delay which grows quadratically with the number of communication hops. 
To reduce this delay, we propose a semi-decentralized approach. This approach uses multiple cloudlet devices each handling a subgraph of the transportation graph. A cloudlet is a moderate computing capability device placed at a base station (BS) and able to communicate with the taxis in its coverage area. From now on, we refer to the cloudlet with its network as a cloudlet network (CLN). As for the BS, we assume a 5G small cell such as the architecture of eNodeB BS detailed in \cite{ge2017energy}. The contributions of this paper can be summarized as follows.

\begin{itemize}[leftmargin=*]
\item We consider taxi demand and supply forecasting at a taxi level. Predicting on a taxi-node level provides drivers with immediate information on the availability of pick-ups and supply of other taxis in a region surrounding them.
\item We propose a novel heterogeneous graph-based algorithm for taxi demand and supply prediction utilizing hetGNNs. We model the transportation system as a heterogeneous graph of taxis as nodes linked with three relationship types derived from road connectivity, location proximity, and target destination proximity. The proposed algorithm exploits these relationships to improve the prediction.

\item We propose a \textit{semi-decentralized} approach to GNN-based traffic prediction. This approach is proposed to mitigate the scalability of centralized GNN inference and the huge message-passing delay in decentralized GNN inference.

\item We propose an adaptive node-CLN assignment to minimize inter-CLN communication. We develop a heuristic protocol for this assignment in a distributed manner across cloudlet devices.
\end{itemize}

\par Experiments on real-world taxi data show a high-accuracy prediction of the proposed taxi demand and supply forecasting algorithm compared to the state-of-the-art, represented by DCRNN \cite{li2017diffusion}, Graph WaveNet \cite{wu2019graph}, and CCRNN \cite{ye2021coupled} being leading GNN-based approaches. Experiments also show that the inference time delay in decentralized GNN operation grows quadratically with the number of message-passing hops. Also, the proposed semi-decentralized GNN approach is shown to reduce the overall inference time by about 10 times compared to centralized and decentralized inference. The source code and datasets used in this paper are available on \url{https://github.com/mahmoudkanazzal/SemidecentralizedhetGNNLSTM}. 

\par The rest of this work is organized as follows. Section \ref{Section2} revises related works. The proposed taxi demand and supply algorithm and semi-decentralized GNN approach are detailed in Section Section \ref{Section3}. Section \ref{Section4} presents experiments and results, with the conclusions in Section \ref{Section5}.

\section{Related Work}
\label{Section2}

\subsection{DL methods for traffic demand forecasting}

\par Similar to their use in other application areas, DL models achieve performance gains in a variety of traffic forecasting problems. This is due to their ability to leverage dependencies among training data. Along this line, recurrent neural networks such as LSTM \cite{hochreiter1997long} are used to exploit time correlations \cite{niu2019predicting}, whereas convolutional neural networks (CNNs) utilize spatial dependencies \cite{niu2018real}. An LSTM is particularly well-suited for prediction tasks that involve temporal dependency. This is because LSTM networks can remember and use previous information over extended periods. A more recent research trend combines LSTMs and CNNs to exploit spatiotemporal correlations in what is known as ConvLSTM \cite{liu2019contextualized,feng2021using}. However, these DL approaches share a common restriction; they overlook existing and potentially useful relationships across entities in a traffic system, such as taxis, roads, and customers. Such relationships model other types of dependencies such as road connectivity and proximity. For instance, the demands at two railway stations in the same city are very likely to be correlated \cite{xu2019incorporating}, even though the two stations may be distant. 

\subsection{GNN models for traffic demand forecasting}

\par Graph data structures appear naturally in many fields such as user accounts in a social network and vehicles in a traffic system \cite{meng2023trajectory}. GNNs extend DL to graph data by combining graph structure and node information through message passing and aggregation \cite{hamilton2017representation,sahu2017ubiquity}. This combination enables GNNs to produce node embeddings to serve multiple downstream graph tasks such as node classification (inferring the class label of a node), link prediction (estimating the likelihood of a link to exist between given
nodes), and graph classification (inferring a property of the graph as a whole). Each node in a graph has a computational graph composed of its $L$-hop neighboring nodes. Node embeddings are obtained by alternating between message passing, i.e., communicating local information across nodes, and aggregation where received messages along with previous node information are used to obtain an updated node embedding. Message passing is done according to the topology of the graph, whereas aggregation is done by the neural network layers of the GNN model obtained by training over graph data \cite{kiningham2022grip,yan2020hygcn}. 

\par As DL methods overlook useful relationships, recent literature considers modeling the traffic system as a graph and applying a GNN approach to problems such as taxi demand and supply forecasting and flow prediction \cite{liang2023semantics}. A city area is divided into many regions and each region is represented by a node in the graph. Several works assume different edges linking these nodes such as having a common origin-destination relationship between two regions if there is a taxi moving from one region (origin) to the other (destination) \cite{xu2019incorporating}. A more recent example work \cite{davis2020grids} considers dividing a city into non-uniform regions serving as the graph nodes linked with edges representing their road connectivity.

\begin{figure}[!htb]
\centering
\resizebox{0.99\columnwidth}{!}{
\begin{tabular}{cc}
\includegraphics[width=10cm]{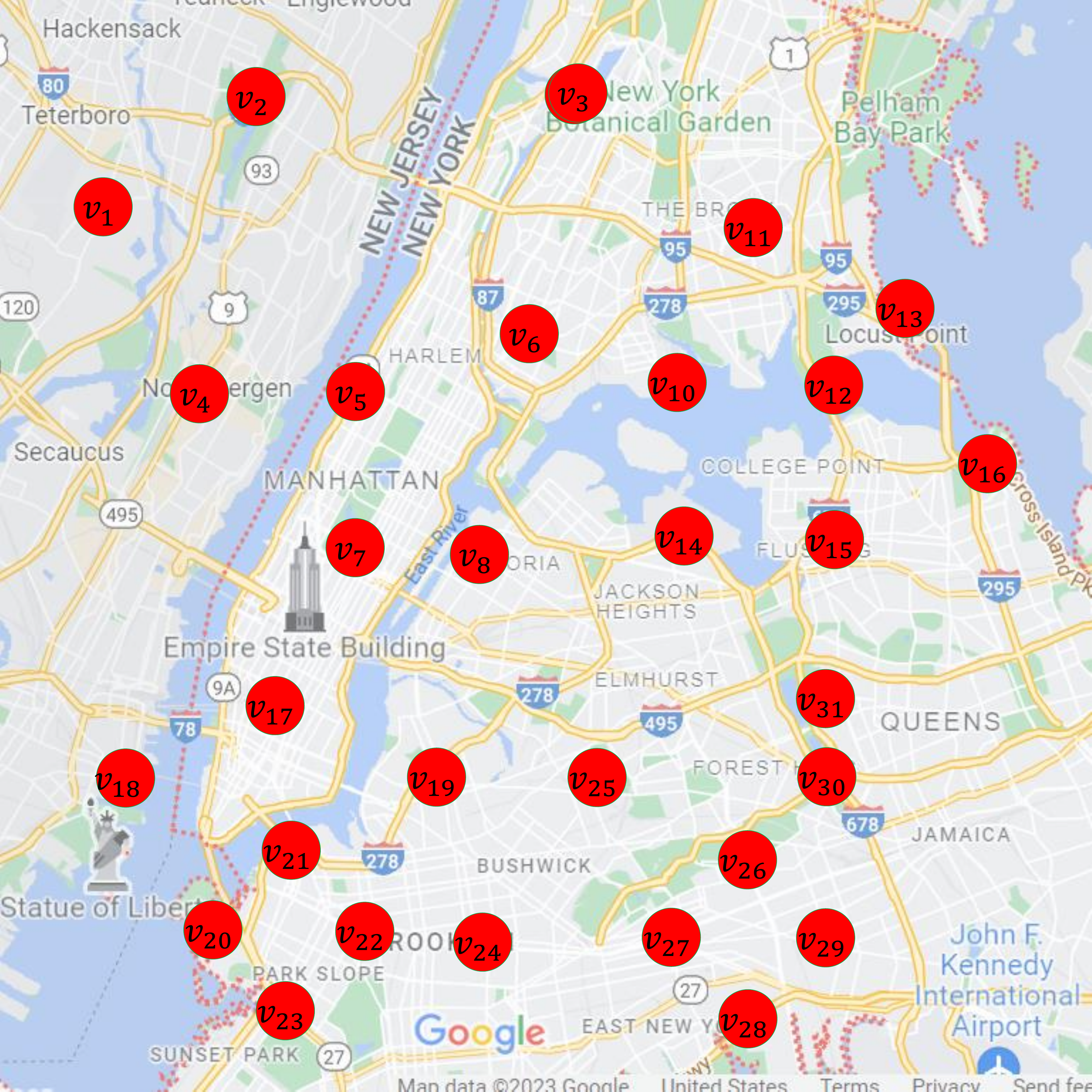}
&
\includegraphics[width=10cm]{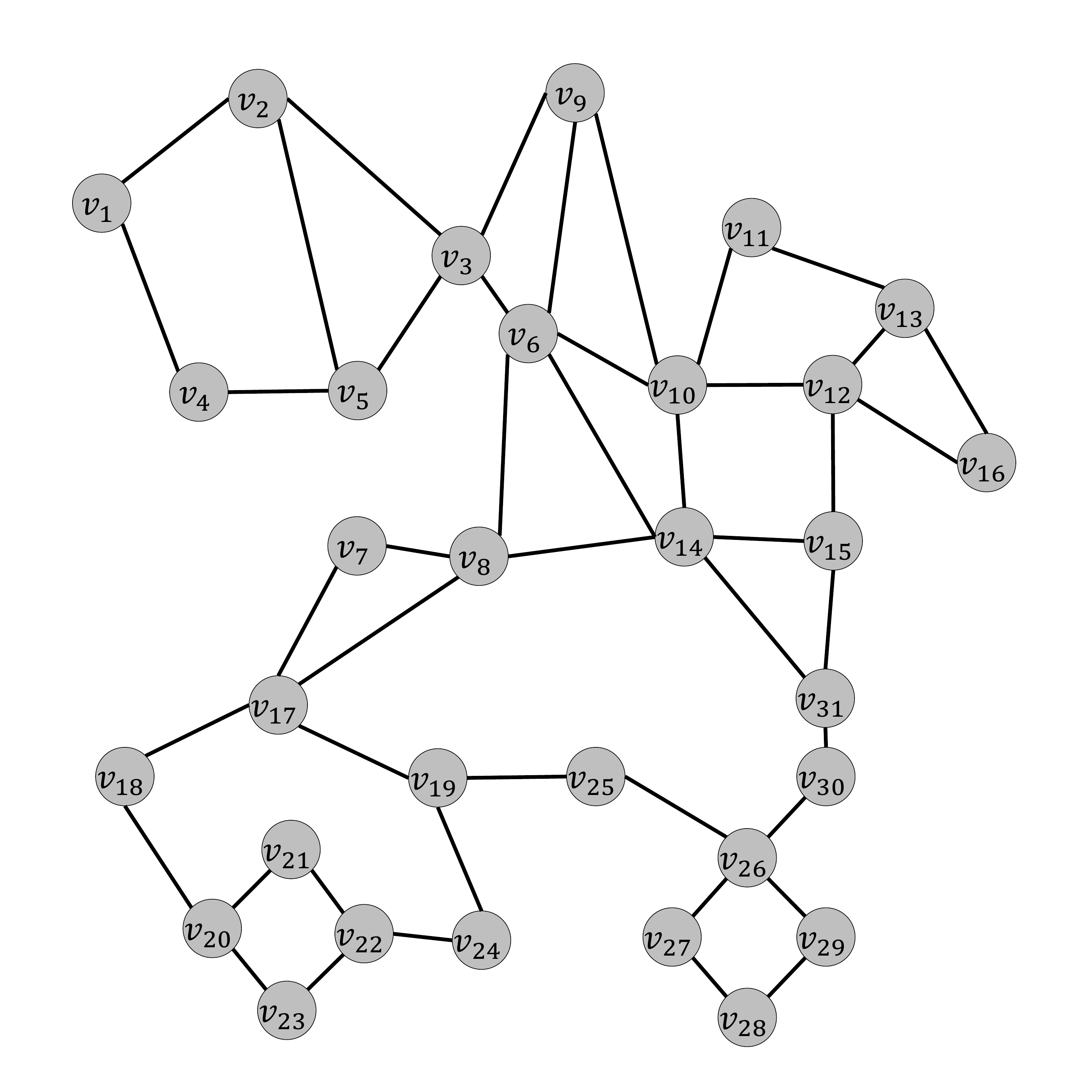}
\\
\Huge{(a)}
&
\Huge{(b)}
\end{tabular}}
\caption{Taxis in a city (a) and their corresponding graph representation (b).}
\label{from_taxi_to_graph}
\end{figure}

\par Despite the clear advantages of GNNs, the existing research body on their usage for taxi demand and supply forecasting assumes homogeneous graphs. So, (homogeneous) GNNs along with other DL models such as LSTM are mainly used to perform the forecasting. Still, when modeling the traffic system as a graph, it may contain different types of nodes with different relation types. So, restricting the use of GNNs to homogeneous models overlooks this richness of relation types and limits the potential of GNNs. Therefore, it is advantageous to model traffic systems as HINs processed by hetGNNs \cite{zhang2019heterogeneous} where the prediction can be improved by incorporating multiple relationships between the graph nodes. The main challenge hetGNNs face is handling heterogeneity. While some primitive hetGNNs project the HIN on the graph space to eliminate its heterogeneity \cite{hu2020heterogeneous}, others use the metapath concept \cite{sun2011pathsim} \footnote{A meta path is a composition of relations linking two nodes.} to maintain and utilize the heterogeneity \cite{guan2022personalized,wang2021self,zheng2021multi}. This is achieved by decomposing the HIN into multiple metapaths encoded to get node representations under graph semantics. hetGNNs have been shown to outperform their GNN ancestors in many applications such as malicious payment detection \cite{liu2018heterogeneous}, drug/illegal trade detection \cite{qian2021distilling}, and network intrusion attack detection \cite{pujol2022unveiling}.

\subsection{Decentralizing GNN inference}
\par Training and testing of GNN models over large graphs require huge memory and processing costs. This is because graph nodes are mutually dependent. Thus, they can not be arbitrarily divided into smaller subgraphs. Techniques such as {neighborhood sampling} \cite{hamilton2017inductive} may ease the problem to some extent. Still, even a sampled computational graph and the associated features may not fit in the memory of a single device \cite{zheng2020distdgl}. Thus, centralized GNN operation faces a scalability limitation \cite{zheng2020distdgl,zeng2022gnn}. 

\par To mitigate the scalability limitation of centralized GNNs, decentralized (peer-to-peer) node inference 
has been applied to a few GNN applications like robot path control \cite{li2020graph,tolstaya2020learning} and resource optimization in wireless communication \cite{lee2021decentralized}. Decentralization naturally promotes GNNs' scalability. Still, it requires excessive communication overhead between nodes \cite{zheng2020distdgl}. In turn, this communication delay significantly slows down the operation of a GNN because the progression of calculations across GNN layers needs to wait for a 2-way message passing to deliver messages from $L$-hop neighbors. Another disadvantage of decentralized inference is the difficulty of coordinating and synchronizing the operation of all nodes. Another less significant disadvantage is the need for each node to maintain and operate a copy of the GNN model.

\par While the literature considers either centralized or decentralized GNNs, it came to our knowledge while developing this work that there is another work \cite{zeng2022gnn} on distributed GNN operation. \cite{zeng2022gnn} proposes adaptive node to cloud server assignment minimizing a general cost function with the servers of different computational powers. However, this is done in a centralized manner; a solver needs to know the graphs of nodes and cloud servers to do the assignment. In our work, we optimize the assignment in a distributed manner at the cloudlet level. Also, according to \cite{zeng2022gnn}, any node may be assigned to any cloud server, while our work focuses on the boundary nodes at each CLN and assigns them to their CLN or an adjacent one taking the geometry into account. Also, while our work focuses on minimizing the communication delay, \cite{zeng2022gnn} adopts a general cost function where the challenge is mapping nodes to cloud servers of varying computational capabilities while optimizing the other costs including the delay. It is also noted that \cite{zeng2022gnn} does not compare centralized and decentralized GNN implementations or consider their trade-offs.

\begin{figure*}[!hbt]
\centering
\resizebox{0.9\textwidth}{!}{
\begin{tabular}{ccc}
\includegraphics[width=10cm]{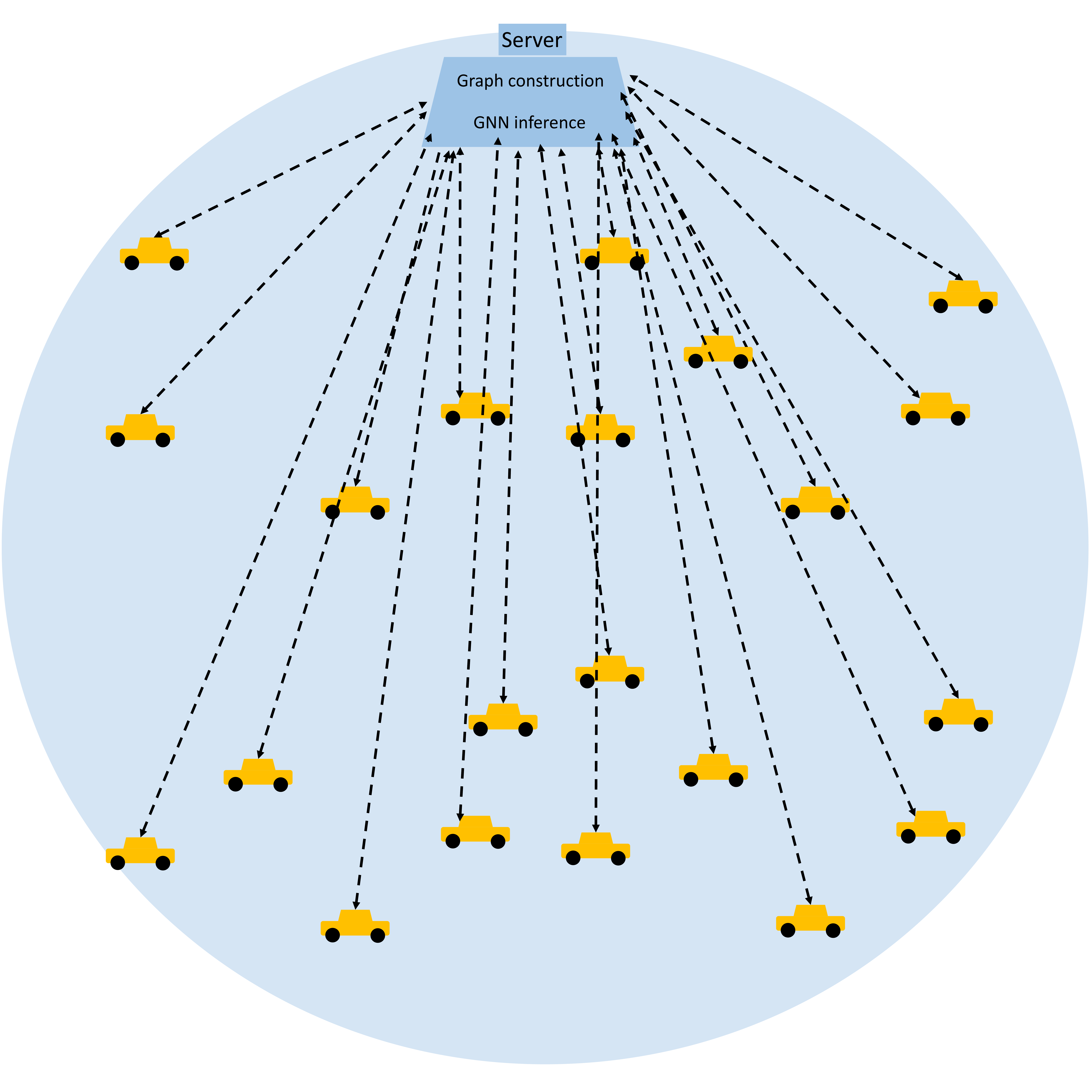}
&
\includegraphics[width=10cm]{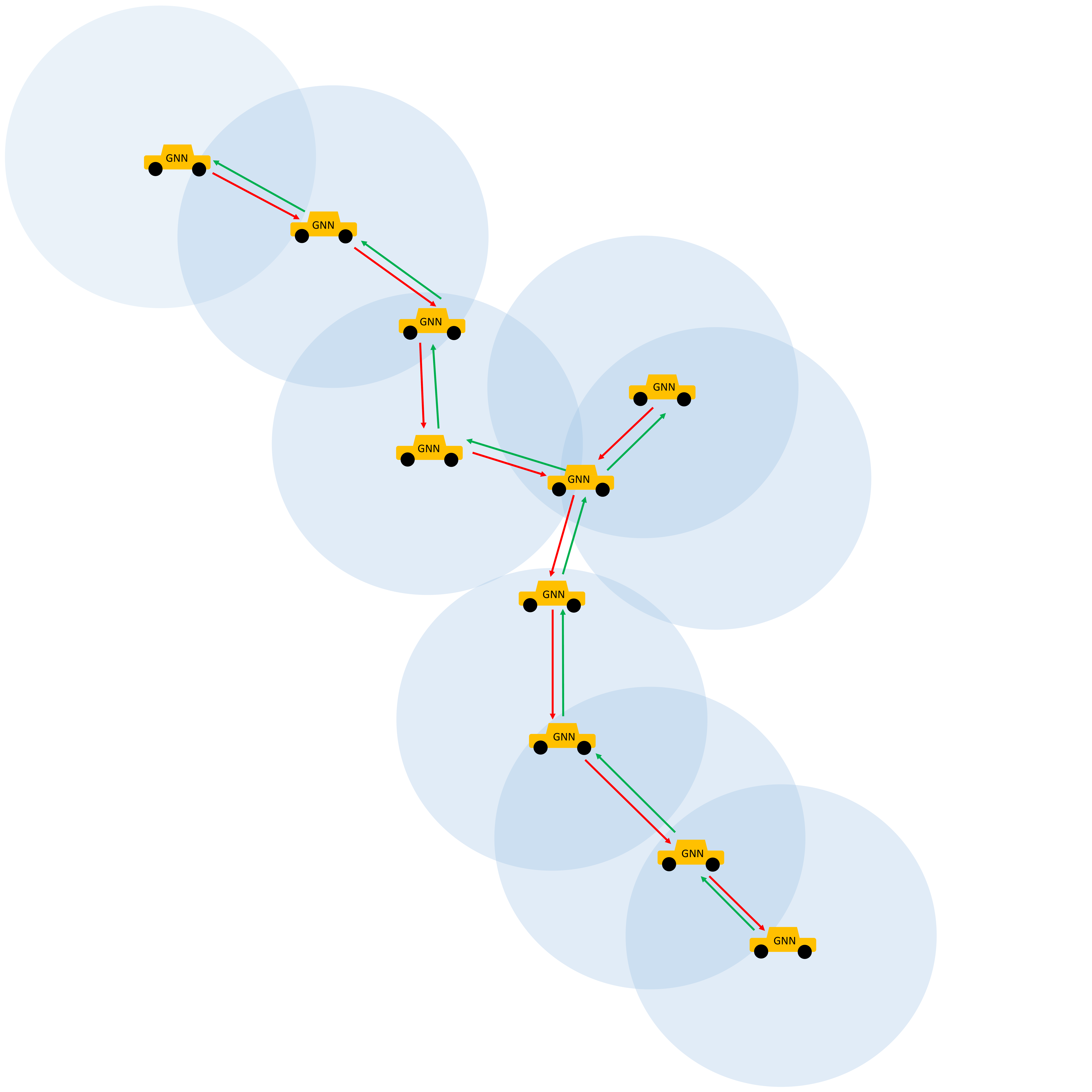}
&
\includegraphics[width=10cm]{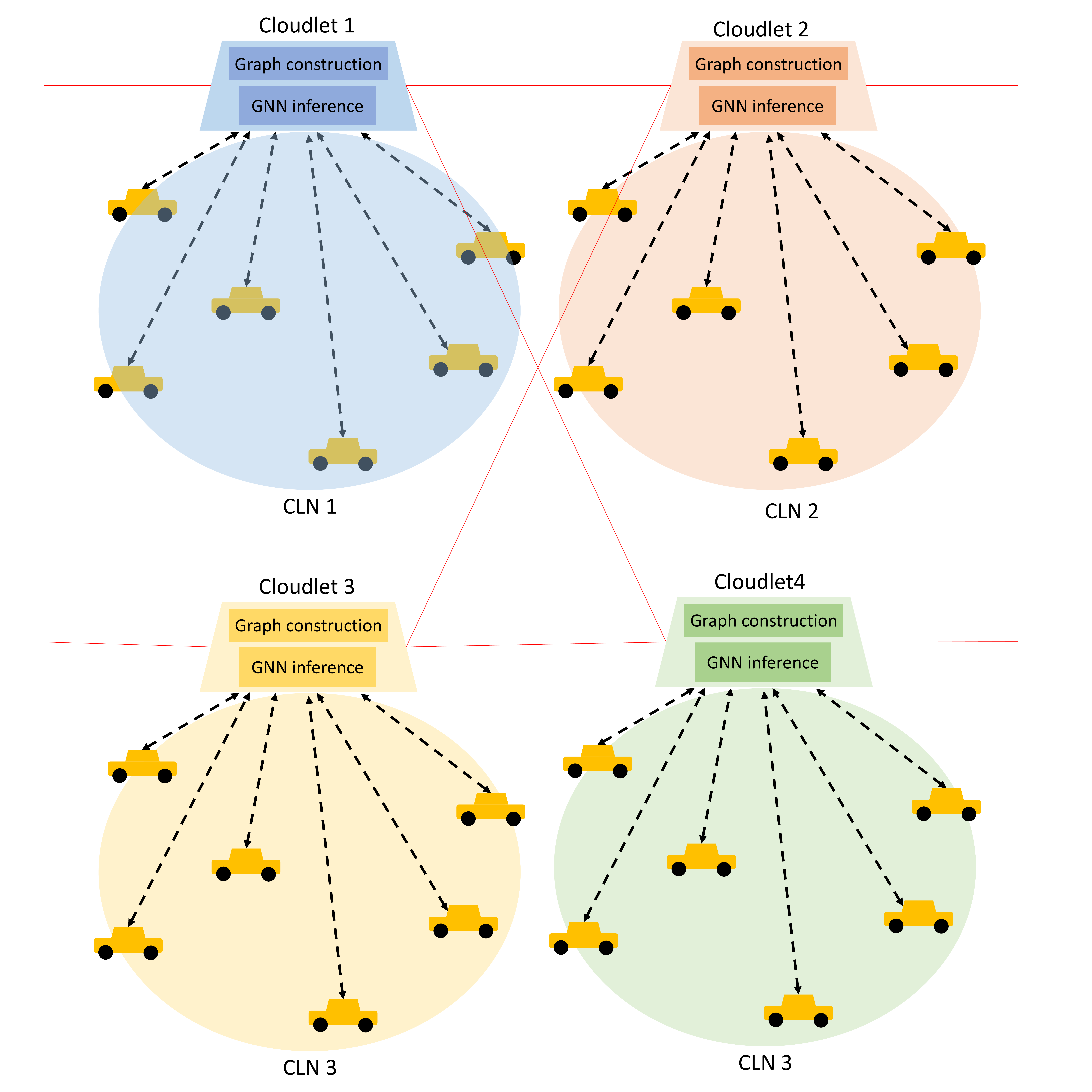}
\\
\Huge{(a)}
&
\Huge{(b)}
&
\Huge{(c)}
\end{tabular}}
\caption{Three possible computation settings. (a) centralized: taxis upload their messages to a central server that performs the computations and returns the results to the taxis. (b) decentralized: taxis exchange messages with their neighbors and perform computations locally. (c) semi-decentralized: taxis in a CLN upload their messages to a cloudlet. The cloudlet performs the computations and returns them to taxis.}
\label{threesettings} 
\end{figure*}

\section{The proposed work}
\label{Section3}

\subsection{System model}
\par The system model considered in this paper is a taxi service system composed of many taxis operating in a certain region/city as shown in Fig. \ref{from_taxi_to_graph}-a. We set the objective of the system as providing future predictions for the demand (e.g., passengers) and supply (e.g., vacant taxis) in the vicinity around each taxi represented by the red circles in this figure. The approach assumed in this work is a GNN approach based on a graph representation of the taxis as shown in Fig. \ref{from_taxi_to_graph}-b. To this end, we study and compare the following approaches to GNN inference with the taxi graph.
\begin{itemize}[leftmargin=*]
\item A fully centralized approach: as represented in Fig. \ref{threesettings}-a, a server or cloud is placed at a BS with a communication range converging the entire operation area. Taxis upload their local messages to the server which also keeps track of their locations. The server uses this information to obtain nodes' computational graphs and uses a local GNN to obtain updated messages. According to the graph structure, the server performs message passing by computation instead of communication. Next, the server sends the updated node messages to their taxi nodes. For taxi-server communication, a communication network under the ITS-G5 standard \cite{mannoni2019comparison} is assumed.

\item Our fully decentralized approach: taxis have GNN models locally and can only communicate with the taxis in their network's coverage area, as represented in Fig. \ref{threesettings}-b. This way, each taxi forms its computational graph. Taxi-to-taxi communication is done through a wireless ad-hoc network such as the one in \cite{miya2021experimental}.

\item Our proposed \textit{semi-decentralized} approach as shown in Fig. \ref{threesettings}-c: This approach uses cloudlets centered at CLNs. Taxis in the coverage area of each CLN use it to send their messages to its cloudlet device. Similar to the centralized setting, the cloudlet uses the uploaded taxi messages along with their locations to obtain node computational graphs and uses a local GNN to obtain updated messages. Message passing for the edges in the CLN is done by computation instead of communication as represented by the dashed lines in Fig. \ref{threesettings}-c. However, the messages between connected taxis in adjacent cloudlets are shared through cloudlet-cloudlet communication, as represented by the solid lines in Fig. \ref{threesettings}-c.
\end{itemize}

\begin{figure*}[!bh]
\centering
\resizebox{0.96\textwidth}{!}{
\begin{tabular}{cc}
\includegraphics[width=7cm]{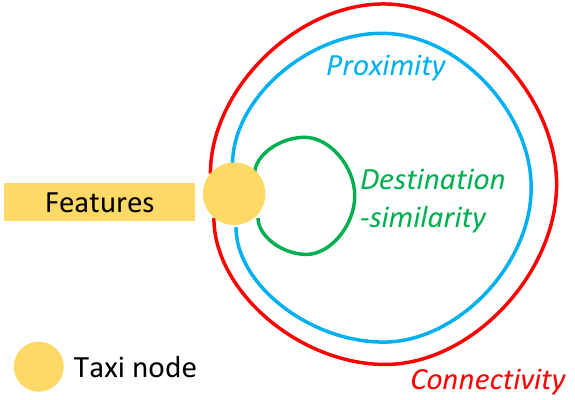}& 
\includegraphics[width=12cm]{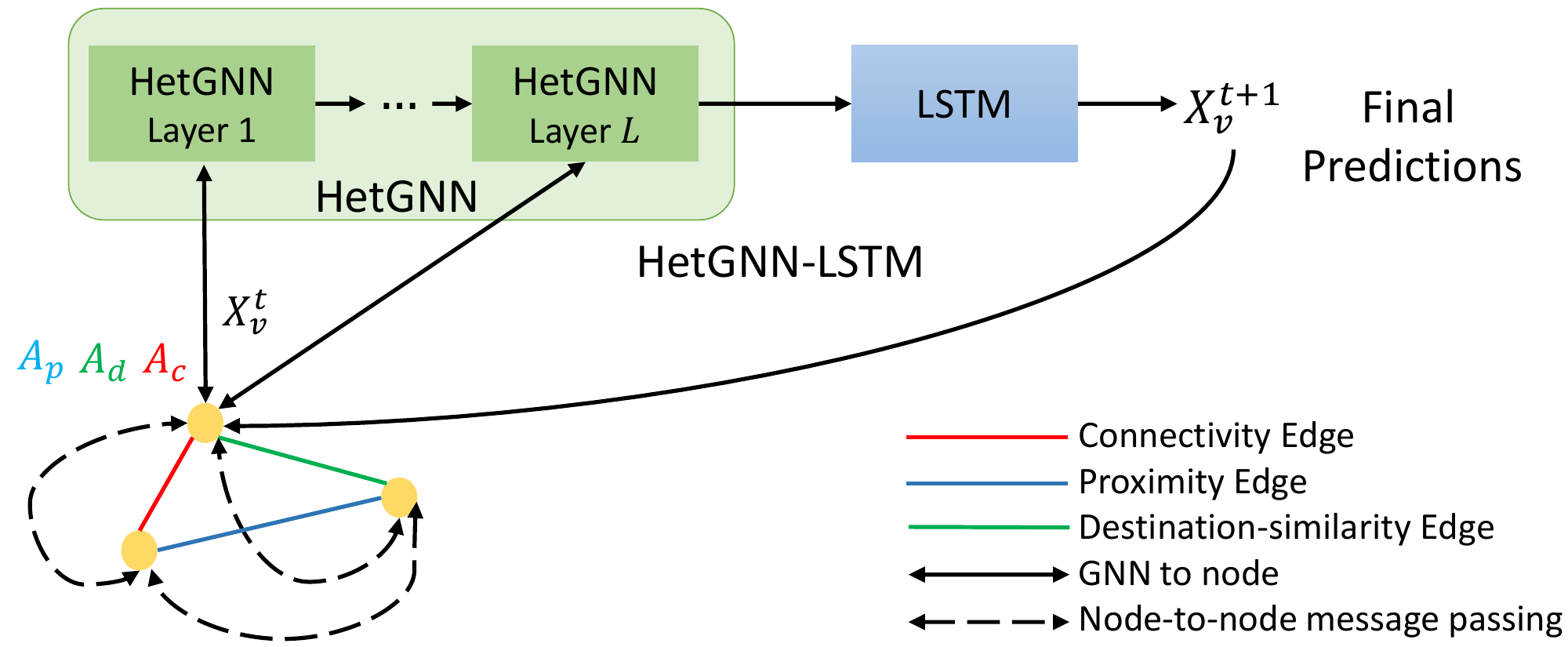}\\
{(a)} & {(b)}
\end{tabular}}
\caption{The network schema of the proposed HIN in (a) and the proposed system architecture in (b).}
\label{architecture} 
\end{figure*}

\subsection{Graph construction and problem formulation}

\par We represent the transportation system as a HIN composed of taxis as its nodes linked with edges of three types. First, is a \textit{connectivity} edge \cite{davis2020grids} representing road connectivity. Second, is a \textit{proximity} edge \cite{xu2022application} linking nearby taxis. Third, we define a \textit{destination-similarity} edge linking taxis going to nearby destinations. Accordingly, a \textit{connectivity} adjacency matrix is as 
 follows \cite{davis2020grids}.
\begin{equation}
\boldsymbol{A}_c[i,j]= \begin{cases}1 & \text { if there is a road connecting nodes } i \text { and } j, \\ 0 & \text { otherwise. }\end{cases} 
\end{equation}

\noindent A \textit{proximity} adjacency matrix is as follows \cite{xu2022application}.
\begin{equation}
\boldsymbol{A}_p[i,j]= \begin{cases}\dist(p_i, p_j) & \text {if } \dist(p_i, p_j)<th_p, \\ 0 & \text { otherwise. }\end{cases} 
\end{equation}
\noindent where $\dist(p_1, p_2)$ is a function of the Euclidean distance between taxi positions $p_1$ and $p_2$, and $th_p$ is a certain threshold. Next, our proposed \textit{destination-similarity} adjacency matrix is as follows.
\begin{equation}
\boldsymbol{A}_d[i,j]= \begin{cases}\dist(d_i, d_2) & \text{if } \dist(d_i, d_2)<th_d, \\ 0 & \text { otherwise. }\end{cases} 
\end{equation}
\noindent where $\dist(d_1, d_2)$ is a measure of the Euclidean distance between destinations $d_1$ and $d_2$, and $th_d$ is a prescribed threshold. 

\par The structure of the proposed HIN is represented in the network schema in Fig. \ref{architecture}-a. We denote the HIN at a  time instant $t$ by $G^t=(\mathcal{V}^t, E_c^t, E_p^t, E_d^t)$, or equivalently, $G^t=(\mathcal{V}^t, \boldsymbol{A}_c^t, \boldsymbol{A}_p^t, \boldsymbol{A}_d^t)$ where $\mathcal{V}^t$ is the node-set, $E_c^t, E_p^t,$ and $ E_d^t$ represent the connectivity, proximity, and destination-similarity edges, respectively, $\boldsymbol{A}_c^t, \boldsymbol{A}_p^t,$ and $ \boldsymbol{A}_d^t$ denote the connectivity, proximity, and destination-similarity adjacency matrices, respectively. For simplicity, we represent the operation of the system on a time slot basis and assume the graph is fixed during a time slot. At a time step $t$, each taxi knows the $P$-step historical demand and supply values of its current region of dimensions $m \times n$ taxi positions, which serve as the node message (attributes). 

\par The objective at each node is to predict the demand and supply values for the next $Q$ time slots. In this sense, each taxi driver will be informed on both the availability of future passengers and other vacant taxis in an $m \times n$ vicinity around their taxi. At a time instant $t$, the graph $G^t$ has an overall feature matrix $\mathbf{X}_t \in \mathbb{R}^{N \times d}$ where $d= m\times n$ is the input feature dimension and $N^t$ is the number of nodes. So, we intend to obtain a mapping function $\mathcal{F}$ as follows. 
\begin{equation}
\left[\mathbf{X}_{t-P+1: t}, G^t\right] \stackrel{\mathcal{F}}{\longrightarrow} \mathbf{X}_{t+1: t+Q},
\end{equation}
\noindent where $\mathbf{X}_{t+1: t+Q} \in \mathbb{R}^{Q \times N^t \times d}$ and $\mathbf{X}_{t-P+1: t} \in \mathbb{R}^{P \times N^t \times d}$.

\subsection{A hetGNN-LSTM algorithm for taxi demand and supply prediction in a semi-decentralized approach} 

\subsubsection{The proposed hetGNN-LSTM algorithm}

\par To incorporate multiple edge types and time dependency in the prediction, we propose a hetGNN-LSTM algorithm as described in Fig. \ref{architecture}-b. For simplicity, we first present its operation in a centralized setting and then in a decentralized setting. After discussing the shortcomings of the centralized and decentralized settings, we present its use in the proposed semi-decentralized setting. In the centralized approach, first, the server constructs a HIN according to the network schema in Fig. \ref{architecture}-a. For each node, messages are shared across the HIN; then the hetGNN layer outcomes are calculated accordingly, and the process continues. Eventually, the final node embedding is obtained after $L$-hop messages are exchanged from the neighbors to the node in question. After that, these embeddings are fed to an LSTM model to produce the eventual demand predictions. The predictions are then sent to their respective nodes. However, this centralized operation lacks scalability due to the huge amount of computation done at the server. This suggests decentralizing the operation.

\par In the decentralized setting, represented in Fig. \ref{threesettings}-b, each taxi maintains a copy of the hetGNN-LSTM model and it is assumed to do two main tasks. The first task is establishing the connection to receive the messages shared from its $L$-hop taxi nodes and using them along with its local information to obtain its final predictions, whereas the second task is sending its messages to these neighbors so that they can operate their GNNs. Due to the absence of a central server, a node needs to identify its $L$-hop neighbors and communicate with them.  However, this is restricted to the communication abilities of these nodes and may not fully make use of the HIN structure between distant nodes. More importantly, the node-node message passing delay limits the number of achievable communicating hops at a reasonable inference delay. 

\par Intuitively, the delay in decentralized GNN inference mainly depends on the message passing delay which significantly increases with the number of communication hops. To quantify this dependency, in Theorem 1, we derive approximate bounds for the overall inference time delay in decentralized GNNs. We show that this delay increases quadratically with the number of communication hops. Furthermore, this increase is determined by the topology of the computational graph of a node. The lower and upper bounds of GNN inference delay are determined mainly by the maximal degree\footnote{A node's degree is the number of edges connected to it \cite{ao2012bounds}.} of nodes in each communication hop, and the summation of node degrees in each communication hop, respectively.

\begin{thm}
In decentralized GNN inference, the overall inference delay with $L$-hop message passing, denoted by $\Delta$, is within the following topology-dependent bounds with quadratic growth with $L$.
\begin{multline}
2t_r\sum_{l=1}^L(L-l+1)[ld_i+\max_{ x\in N_l(i)} \{d_x\}+(l+1)t_p ]\\
\leq 
\Delta \leq \\
2t_r\sum_{l=1}^L(L-l+1)[ld_i+\sum_{x \in N_l(i)} d_x+(l+1)t_p]
\end{multline}
\noindent where $L$ is the number of message passing hops (equivalently, the number of GNN layers), $t_r$ is the packet transmission delay of the wireless network used for message passing, $d_i$ is the degree of node $i$, $N_l(i)$ is the set of $l$-hop-away neighbors to node $i$, and $t_p$ is the GNN layer processing delay. 
\end{thm}

\par
\begin{figure}[!htb]
\centering
\resizebox{0.99\columnwidth}{!}{
\includegraphics[width=12cm]{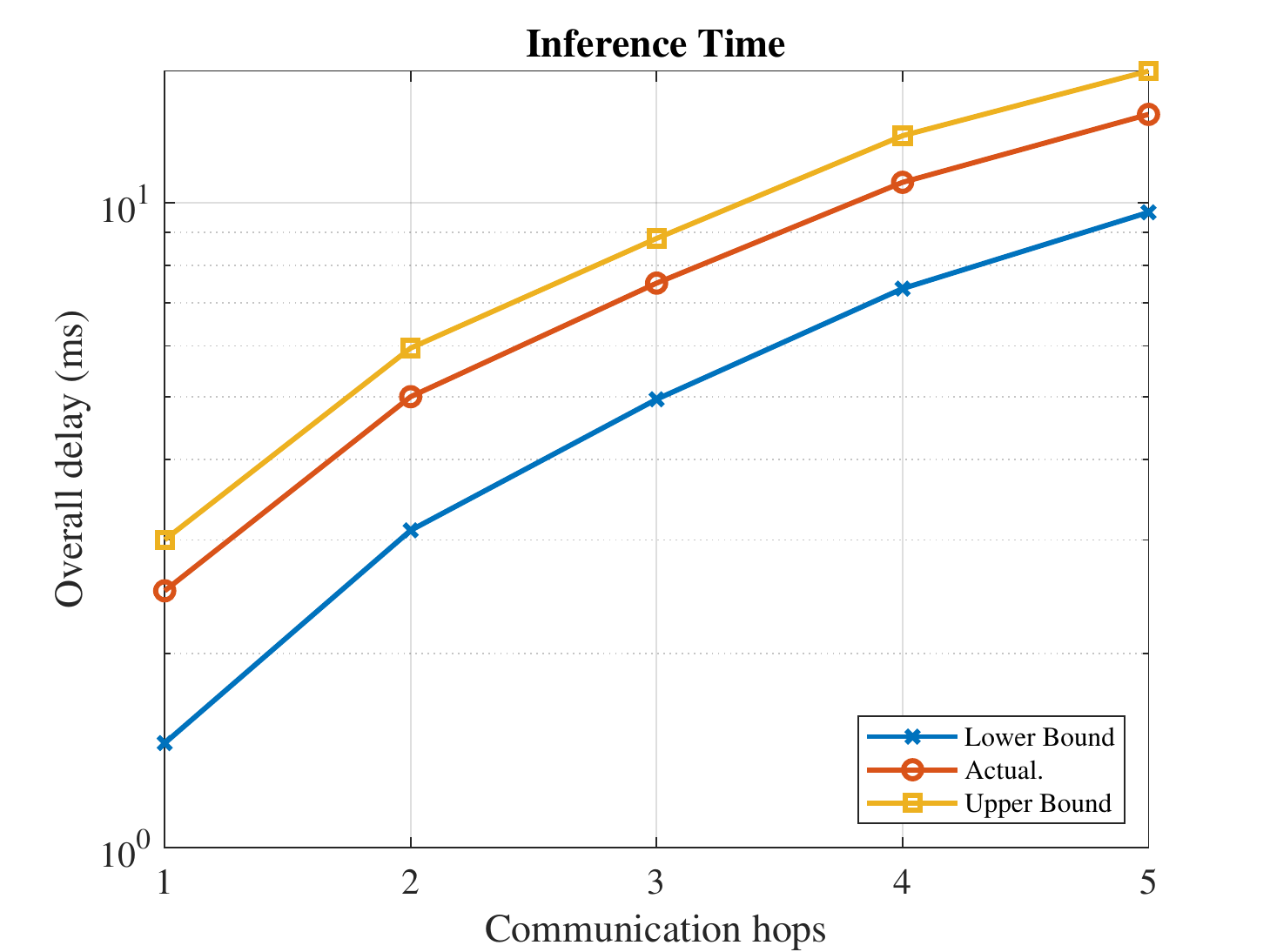}}
\caption{Actual values of overall GNN inference delay versus the number of communication hops and their corresponding bounds of Theorem 1.}
\label{bound_comparison} 
\end{figure}

 The proof of Theorem 1 is in the Appendix. To investigate the relationship between the number of hops and the overall GNN inference delay in real-world scenarios, we present the following experiment. A total of 255 taxis in a city region are considered to work in a decentralized GNN setting. We assume an ad hoc wireless network connecting taxi nodes. We calculate the overall inference time as specified in Section \ref{Section4}. For $L$ hop values ranging from 1 to 5, an $L$-hop computational graph of a node is obtained, and the overall message passing delay is calculated. We also calculate the GNN inference delay bounds presented in Theorem 1. We repeat this experiment for 10 trials and plot average values of the overall GNN inference delay and its bounds versus hops in Fig. \ref{bound_comparison}. As shown in this figure, the overall inference delay grows with increasing communication hops at a quadratic proportionality. Also, the actual delay is within the bounds specified in Theorem 1. This result shows the message passing delay bottleneck in decentralized GNN inference.

\begin{figure*}[!t]
\centering
\resizebox{0.99\textwidth}{!}{
\begin{tabular}{ccc}
{\includegraphics[width=42cm]{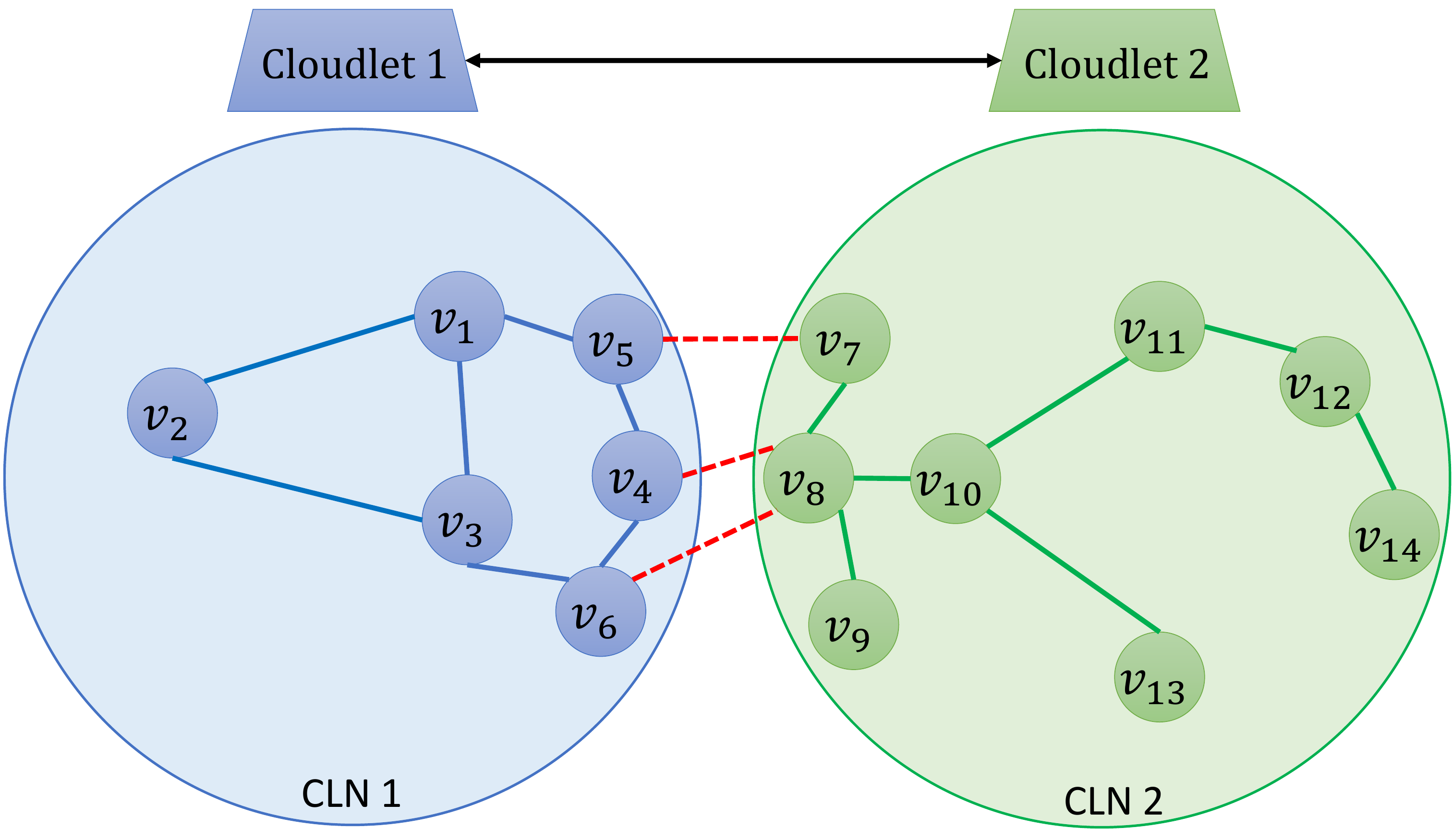}}
&
{\includegraphics[width=42cm]{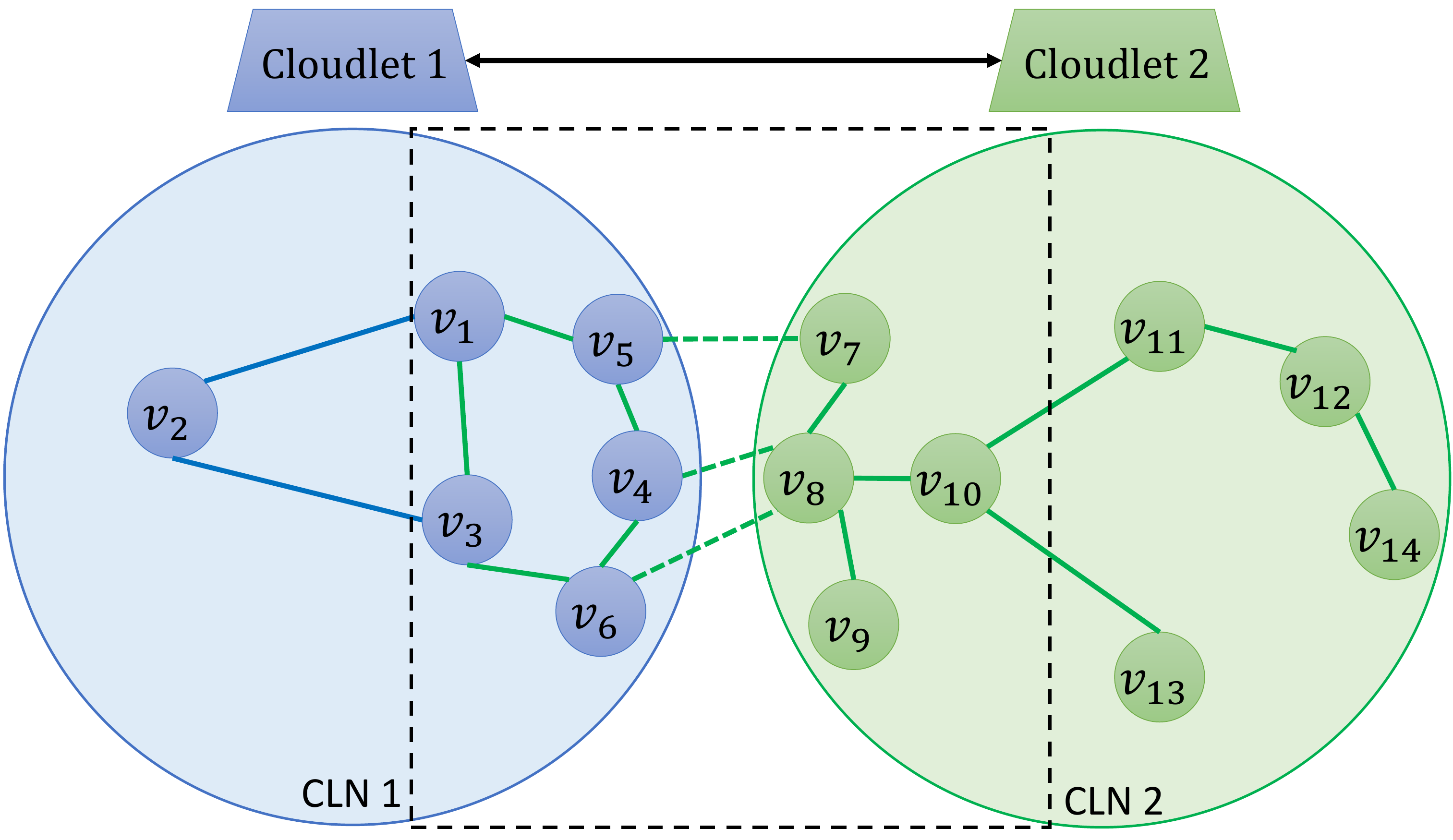}}
&
{\includegraphics[width=42cm]{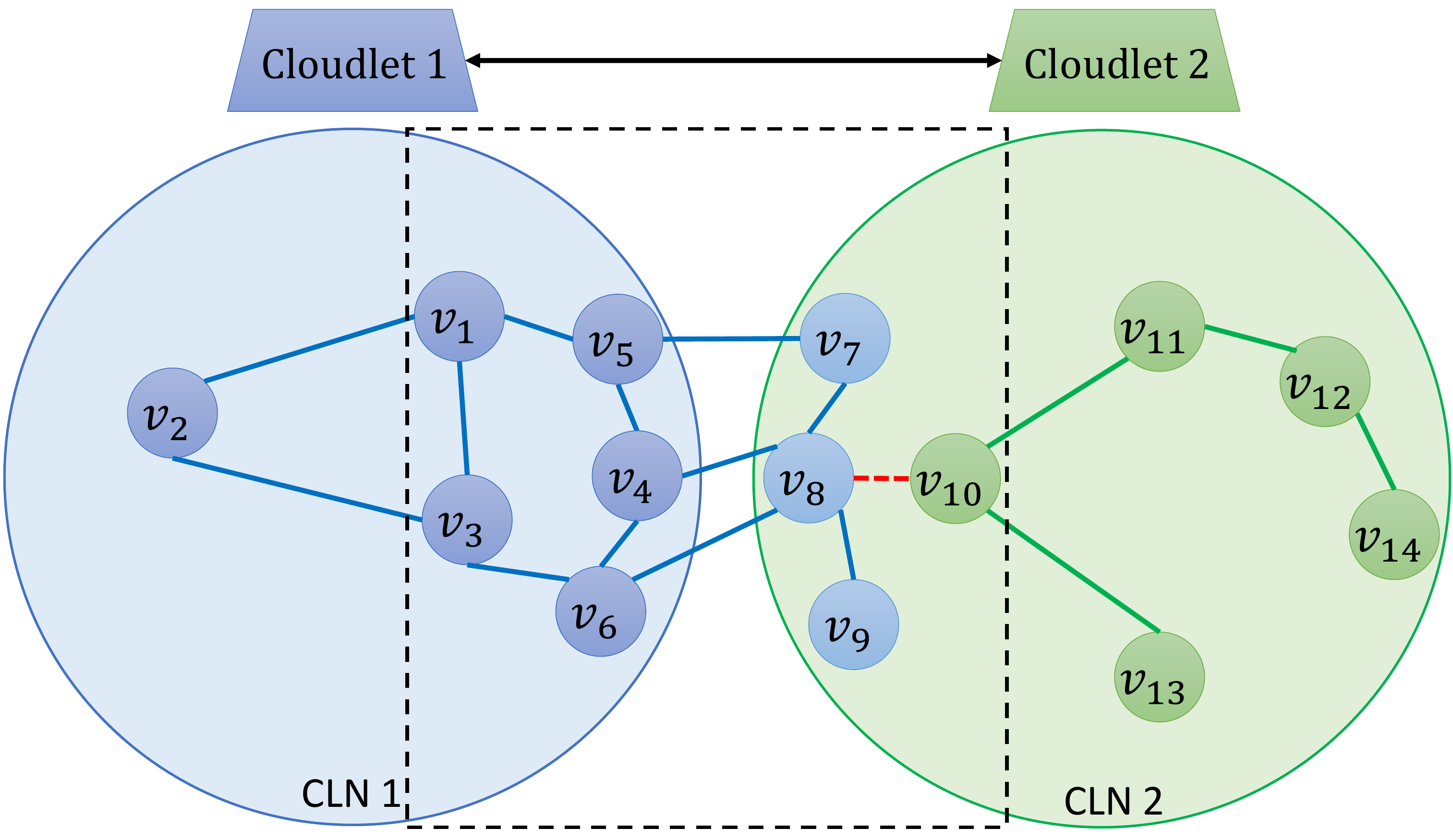}}
\\
\fontsize{90}{90}\selectfont{(a)}
&
\fontsize{90}{90}\selectfont{(b)}
&
\fontsize{90}{90}\selectfont{(c)}
\end{tabular}}
\caption{(a) Coverage-based assignment with 3 inter-CLN edges, (b) the shared boundary subgraph to be partitioned, (c) adaptive assignment with 1 inter-CLN edge.}
\label{partitioning}
\end{figure*}

\par To resolve the limitations of scalability and excessive message passing delay in the centralized and decentralized schemes, respectively, we propose a semi-decentralized approach. As shown in Fig. \ref{threesettings}-c, this approach uses a set of CLNs that span the work area of the taxis. Each CLN is associated with a certain city area and thus establishes a sub-HIN accordingly. Then, taxis in each CLN's region (sub-HIN) communicate their messages to the cloudlet, which performs predictions using its copy of the hetGNN-LSTM model, and then it communicates these predictions back to their respective taxis. It is noted that some boundary taxi nodes may have edges connecting them to taxi nodes in an adjacent CLN. In this case, the CLNs serving these taxis need to exchange messages about these nodes for each GNN level. Adjacent CLNs exchange information on their boundary taxis that have edges across these CLNs, as represented by the solid lines in Fig. \ref{threesettings}-c. The main steps of the operation in the semi-decentralized approach are outlined in Algorithm \ref{Algorithm1}.

\begin{algorithm}[t!]
\caption{Semi-decentralized GNN inference.}
\label{Algorithm1}
\begin{algorithmic}[1]{
\algsetup{linenosize=\small}
\renewcommand{\algorithmicrequire}{\textbf{Input:}} 
\renewcommand{\algorithmicensure}{\textbf{Output:}}
\REQUIRE Initial CLN region boundaries for each cloudlet, a trained GNN-LSTM model at each cloudlet.
\ENSURE The next $Q$-future predictions of taxi demand and supply in an $m \times n$ vicinity region around each taxi node
\STATE{Each cloudlet $u$ senses the existing $n_u$ taxis in its CLN region.}
\STATE{Each cloudlet $u$ obtains a sub-HIN $G_u^t:(\boldsymbol{A_{u,c}^t}, \boldsymbol{A_{u,p}^t}, \boldsymbol{A_{u,d}^t} \in \mathbb{R}^{n_u \times n_u})$ based on the taxis in its CLN region.}
\STATE{Each cloudlet $u$ obtains the feature matrix of the nodes in its CLN its nodes $\boldsymbol{X_u^t} \in \mathbb{R}^{n_u \times k^t})$}
\STATE{Each cloudlet $u$ determines the boundary nodes in its CLN connected to nodes in an adjacent cloudlet $v$.}
\STATE{Each cloudlet $u$ computes the messages passed across its nodes.}
\STATE{Each cloudlet $u$ sends the messages of its boundary nodes to the CLNs containing their connected nodes.}
\STATE{Each cloudlet $u$ receives the messages from the nodes connected to its boundary nodes through their CLN.}
\STATE{The updated node messages are aggregated and fed to the $l$-th GNN layer to produce new messages.}
\STATE{Each cloudlet $u$ sends the eventual embeddings to their respective nodes in the CLN.}
}
\end{algorithmic}
\end{algorithm}

\subsubsection{Adaptive node-CLN assignment}

\par In semi-decentralized operation, inter-CLN communication needs to happen for boundary nodes at each GNN level. The way the nodes are assigned to CLNs governs the number of inter-CLN edges. Therefore, an adaptive assignment can be employed to minimize the number of inter-CLN edges. We propose using minimum edge-cut graph partitioning to divide the shared subgraph between each pair of adjacency CLNs to serve this goal. As an example, in Fig. \ref{partitioning}-a, uniformly assigning the nodes to CLNs results in 3 edges across the CLNs (denoted by the red dashed lines). However, applying minimum-cut graph partitioning to the nodes in the shared boundary region ($\mathcal{V}_b$) enclosed by the dashed lines in Fig. \ref{partitioning}-b, a new assignment may add nodes $v_7$, $v_8$, and $v_9$ to CLN 1, as seen in Fig. \ref{partitioning}-c. This means that inter-CLN message passing needs to happen only across one node pair ($v_8$-$v_{10}$). This partitioning has to be done in a distributed manner between the CLNs. To achieve this, CLN 1 and CLN 2 share information on nodes and their positions in the boundary subgraph $\mathcal{V}_b$. Nodes in this 2-$L$ region are then to be partitioned between the two CLNs in a distributed 2-way manner. This means that one CLN will do the partitioning and instruct the other. For example, let us assume it is CLN 2. So, the partitioning problem is to optimize an assignment operator $\Psi$ that assigns each node in the shared boundary subgraph ($\mathcal{V}_b$) to belong either to the set of nodes assigned to CLN 1 ($\mathcal{V}^1$) or to that of CLN 2 ($\mathcal{V}^2$), minimizing inter-set edges. The node assignment problem can be 
 formulated as follows.
\begin{equation}
\begin{array}{ll@{}l}
\operatorname*{argmin}_\Psi & \sum_u \sum_v \mathds{1}\{\mathcal{V}^1_u,\mathcal{V}^2_v \} \\
\text{subject to} & (\mathcal{V}^1,\mathcal{V}^2)=\Psi(\mathcal{V}_b) \\
& \mathcal{V}^1 \cap \mathcal{V}^2 =\varnothing
\end{array},\label{eq0}
\end{equation}
\noindent where the indicator function $\mathds{1}\{\mathcal{V}^1_u,\mathcal{V}^2_v \}$ is 1 if there is an edge between the $i$-th node in $\mathcal{V}^1$ and the $j-th$ node in $\mathcal{V}^2$, and is 0 otherwise.

\par The node assignment problem in (\ref{eq0}) can be solved as a minimum edge-cut graph partitioning process. The overall graph is divided into multiple subgraphs with minimum edge cuts, and each subgraph is assigned to a CLN. The graph partitioning problem is known to be NP-complete \cite{karypis1998fast}. However, multi-level graph partitioning as in the METIS \cite{karypis1997metis} algorithm provides suitable solutions by operating in three stages; coarsening the graph into smaller graphs, bisecting the smallest graph, and gradually projecting nodes on each graph partition. Still, the nodes in the taxi graph are aligned in a two-dimensional space. This eases the problem of graph partitioning by restricting partitioning to boundary nodes in neighboring CLNs. So, in this work, we apply k-means clustering between each pair of adjacent CLNs to partition their boundary nodes in a distributed manner. We propose a protocol for achieving adaptive assignment between each pair of adjacent CLNs in a distributed manner in Algorithm \ref{Algorithm2}.

\begin{algorithm}[t!]
\caption{Distributed adaptive node-clouded assignment.}
\label{Algorithm2}
\begin{algorithmic}[1]{
\algsetup{linenosize=\small}
\renewcommand{\algorithmicrequire}{\textbf{Input:}} 
\renewcommand{\algorithmicensure}{\textbf{Output:}}
\REQUIRE Region boundaries for the CLN of each cloudlet.
\ENSURE Adaptive node-CLN assignment maps minimizing inter-CLN taxi communication.
\STATE{Each cloudlet $u$ obtains a sub-HIN $G_u^t:(\boldsymbol{A_{u,c}^t}, \boldsymbol{A_{u,p}^t}, \boldsymbol{A_{u,d}^t} \in \mathbb{R}^{n_u \times n_u})$
}
\STATE{Only the graph topology is important in the following steps
}
\STATE{Cloudlets operate as adjacent pairs and focus on their shared regions}
\STATE{From each neighbor pair, one cloudlet sends the topology of its boundary nodes to its neighboring cloudlet}
\STATE{The cloudlet receiving the shared topology will augment it to its boundary topology}
\STATE{The cloudlet receiving the shared topology uses k-means minimum-cut graph partitioning to divide the boundary area minimizing edge cuts.}
\STATE{The receiving cloudlet will forward the result to its neighboring CLN}
\STATE{Both cloudlets will adjust to the new assignment.}
}
\end{algorithmic}
\end{algorithm}

\subsubsection{Computational complexity analysis}
\par The proposed algorithm assumes taxi nodes and generates predictions in the $m\times n$ vicinity of each node, whereas the existing approaches predict aggregate demand and supply values for each region treated as a node. Assuming a city having $N$ taxis operating in $K$ regions, the proposed algorithm retrieves $2 m\times n\times P \times Q$ data values for $N$ taxis whereas the existing approaches retrieve $2\times P \times Q$ data values for $K$ regions. It is interesting to compare the inference time complexity of the proposed algorithm to that of the existing approaches.

\par The inference time complexity of the existing approaches,  depends on the following factors.
\begin{itemize}[leftmargin=*]
    \item Time complexity of passing 2$PQ$-dimensional messages across $K$ (region) nodes within $L$ GNN layers: assuming a message packet delay $t_r$, this is proportional to $P, Q, K$, and $L^2$ (as concluded in the proof of Theorem 1).
    \item The time complexity of processing 2$PQ$-dimensional messages for $L$ layers per node: assuming a processing delay $\tau$ for each message, this is proportional to $P, Q, L, K$, and $\tau$.
\end{itemize}
\noindent Therefore, the inference time complexity of existing approaches is $O(KPQ(L^2t_r+L \tau)$.
\par The inference time complexity of the proposed algorithm in the semi-decentralized setting is calculated at a cloudlet level since cloudlets operate concurrently. Let us assume that the number of nodes per CLN is $N/K$, on average. The cloudlet inference time depends on the following factors.
\begin{itemize}[leftmargin=*]
    \item Message passing by computation of $mnPQ$ messages for $L$ GNN layers per node. Let us denote by $t_c$ the time to pass an $mnPQ$-dimensional message by communication. So, the time complexity due to this delay is proportional to $N/K, m,n, P, Q,t_c$ and $L^2$.
    \item Layer processing time of $mnPQ$ messages for $L$ layers per node. Again, let us denote this processing time by $\tau$. So, the time complexity due to this delay is proportional to $L\tau$.
    \item Cross-CLN MP for boundary nodes: let us assume a faction $\gamma$ of cloudlet nodes being boundary nodes connected to nodes in other CLNs, and let us denote by $t_{CLN}$ the message packet transmission delay across CLNs. Then, this delay will be $\gamma N/K L^2 t_{CLN}$.
\end{itemize}
\noindent So, the overall time per-cloudlet time complexity is $O(mnPQ t_c N/K+mnPQ N/K L \tau+ \gamma N/K L^2t_{CLN})$.

\section{Experiments and Performance Analysis}
\label{Section4}
\par We present experiments on real-world data to analyze the operation of the proposed hetGNN-LSTM taxi demand and supply forecasting algorithm and the semi-decentralized GNN approach from the perspectives of prediction accuracy and GNN inference delay. 

\subsection{The setup and dataset}
\par The dataset used in this work is adopted from the NYC dataset \cite{taxi2017tlc}. This dataset consists of 35 million taxi trip records in New York City from April 1st, 2016 to June 30th, 2016. For each trip, the following information is recorded; pick-up time, drop-off time, pick-up longitude, pick-up latitude, drop-off longitude, drop-off latitude, and trip distance. As for the hetGNN, we use a heterogeneous graph convolutional network (HeteGCN). We implement the models using PyTorch Geometric (PyG) \cite{fey2019fast} version 2.0.3 built on PyTorch version 1.10.0. Experiments are conducted on a Lambda GPU Workstation with 128 GB of RAM, two GPUs each of 10 GB RAM, and an I9 CPU of 10 cores at a clock speed of 3.70 GHz. Table \ref{table2} lists key hyperparameters of the hetGNN-LSTM model.

\begin{table}
\centering
\caption{Key properties and hyperparameters of the hetGNN-LSTM model used.}
\begin{tabular}{|l|c|}
\hline
Property& Value\\ \hline
Optimizer & Adam \\ \hline
GNN learning rate & 0.005\\ \hline
GNN coefficient decay& 0.001\\ \hline
GNN No. of layers & 3\\ \hline
LSTM learning rate & 0.0001 \\ \hline
LSTMno. of layers & 10 \\ \hline
Input dimensions& 216\\ \hline
Output dimensions & 216\\ \hline
Loss function & Mean squared error (mse) \\ \hline
GNN convolution layer & GCNConv class from PyG \cite{fey2019fast}\\ \hline
\end{tabular}
\label{table2}
\end{table}

\subsection{Performance of the proposed hetGNN-LSTM forecasting algorithm} 

\par In this experiment, we compare the accuracy of taxi demand and supply values predicted by the proposed hetGNN-LSTM algorithm with the following approaches representing the state-of-the-art.
\begin{itemize}[leftmargin=*]
\item DCRNN \cite{li2017diffusion}: 
uses a combination of diffusion convolutional layers and RNNs to make predictions. 
\item Graph WaveNet \cite{wu2019graph}: uses a combination of graph convolutional layers and dilated causal convolutional layers to capture the complex patterns in the data and uses an adaptive adjacency matrix.
\item CCRNN \cite{ye2021coupled}: uses a GCN model with coupled layer-wise graph convolution, which allows for more effective modeling of the spatial-temporal dependencies in transportation demand data with learnable adjacency matrices. 
\end{itemize}

\par Comparisons are conducted in terms of the following quality metrics. 
\begin{itemize}[leftmargin=*]
\item Root mean squared error (RMSE)
$$
\operatorname{RMSE}(\boldsymbol{x}, \hat{\boldsymbol{x}})=\sqrt{\frac{1}{n} \sum_{i}\left(x_i-\hat{x}_i\right)^2}
$$ 
\item Mean absolute percentage error (MAPE)
$$
\operatorname{MAPE}(\boldsymbol{x}, \hat{\boldsymbol{x}})=\frac{1}{n} \sum_{i}\left|\frac{x_i-\hat{x}_i}{x_i}\right|
$$
\item Mean absolute error (MAE)
$$
\operatorname{MAE}(\boldsymbol{x}, \hat{\boldsymbol{x}})=\frac{1}{n} \sum_{i}\left|x_i-\hat{x}_i\right|
$$
\end{itemize}

\noindent where $\boldsymbol{x}=x_1, \cdots, x_n$ denotes the ground truth values of taxi demand and supply, and $\hat{\boldsymbol{x}}=\hat{x}_1, \cdots, \hat{x}_n$ represents their predicted values.

\begin{table}[htb]
\centering
\caption{Performance evaluation of the proposed hetGNN-LSTM algorithm in three decentralization settings, compared to three state-of-the-art approaches in terms of RMSE, MAE, and, MAPE. We provide three comparison cases.}
\resizebox{0.97\columnwidth}{!}{
\begin{tabular}{|l|c|c|c|c|}
\hline
Comparison Case & Method & RMSE& MAE & MAPE (\%) \\ \hline
\multirow{6}{*}{Taxi-region} & DCRNN \cite{li2017diffusion} & 14.79 & 8.43& 21.27 \\ \cline{2-5} 
& Graph WaveNet \cite{wu2019graph} & 13.07 & 8.10& 18.33 \\ \cline{2-5} 
& CCRNN \cite{ye2021coupled}& 9.69& 5.58& 18.36 \\ \cline{2-5} 
& Proposed-SC1 & 14.17 & 11.99 & 10.24 \\ \cline{2-5} 
& Proposed-SC2 & 15.24 & 12.91 & 12.31 \\ \cline{2-5} 
& Proposed-SC3 & 14.29 & 12.24 & 10.87 \\ \hline
\multirow{6}{*}{Region-region} & DCRNN \cite{li2017diffusion} & 14.79 & 8.43& 21.27 \\ \cline{2-5} 
& Graph WaveNet \cite{wu2019graph} & 13.07 & 8.10& 18.33 \\ \cline{2-5} 
& CCRNN \cite{ye2021coupled}& 9.69& 5.58& 18.36 \\ \cline{2-5} 
& Proposed-SC1 & 5.29& 2.46& 7.62\\ \cline{2-5} 
& Proposed-SC2 & 6.72& 3.56& 9.33\\ \cline{2-5} 
& Proposed-SC3 & 5.43& 3.11& 7.96\\ \hline
\multirow{6}{*}{Taxi-taxi} & DCRNN\cite{li2017diffusion} & 21.48 & 20.16 & 26.73 \\ \cline{2-5} 
& Graph WaveNet \cite{wu2019graph} & 19.21 & 18.00 & 24.15 \\ \cline{2-5} 
& CCRNN \cite{ye2021coupled}& 17.16 & 16.84 & 20.61 \\ \cline{2-5} 
& Proposed-SC1 & 14.17 & 11.99 & 10.24 \\ \cline{2-5} 
& Proposed-SC2 & 15.24 & 12.91 & 12.31 \\ \cline{2-5} 
& Proposed-SC3 & 14.29 & 12.24 & 10.87 \\ \hline 
\end{tabular}}
\label{table1}
\end{table}

\par There are two major differences between the operation of the proposed algorithm and the existing algorithms in the literature \cite{li2017diffusion,yu20193d,xu2019incorporating,wu2019graph,davis2020grids,chen2020multi,ye2021coupled,xu2022application}. First, our hetGNN-LSTM algorithm considers dynamically evolving graphs since its node granularity is at the taxi level. Thus, taxis can enter or leave the system generating a dynamic graph. In contrast, the existing approaches model city regions as graph nodes, and therefore, have static graphs. Secondly, our algorithm considers predicting the demands and supplies in a vicinity surrounding each taxi node whereas the existing approaches only predict at the locations of nodes (since each node is a region). Therefore, it is unfair to directly compare our algorithm to the existing approaches as this will compare different quantities. We thus include the following comparison cases where our algorithm is operated on a region level and the other algorithms are operated on a taxi level.

\begin{itemize}[leftmargin=*]
\item A \textit{taxi vs. region} comparison case: in this case, we record the taxi demand and supply values obtained by the proposed algorithm at a 3x3 region surrounding each taxi, and record demand and supply predictions of the baselines at specific city regions. This is an unfair comparison since the proposed algorithm predicts more information compared to the baselines.
\item A \textit{region vs. region} comparison case: from the demand and supply predictions obtained with the proposed algorithm at taxi locations, we calculate the overall predictions made in each city region. We compare these values to the predictions at the same regions obtained by each of the baselines. This is a fair comparison since we compare corresponding predictions.
\item A \textit{taxi vs. taxi} comparison case: we record taxi demands and supplies obtained by the proposed algorithm at taxi locations. For the baselines, we use them to predict demands and supplies at the same taxi locations (rather than on a region level). This is also a fair comparison case since we compare predictions of the same information. 
\end{itemize}

\par For each of the above-mentioned comparison cases, we operate the proposed hetGNN-LSTM model in the three decentralization settings. Namely, centralized, fully decentralized, and semi-decentralized settings denoted by scenarios $SC1$, $SC2$, and $SC3$, respectively. Table \ref{table1} lists the results of this experiment. Considering Table \ref{table1}, let us first compare the performances of the proposed algorithm in centralized, fully-decentralized, and semi-decentralized settings ($SC1$, $SC2$, and $SC3$, respectively). 
The centralized approach $SC1$ has the best performance compared to $SC2$ and $SC3$. This is because it has access to all $L$-hop information for each node. Since $SC3$ uses inter-cloudlet communication to achieve message passing between boundary nodes in adjacent CLNs, its performance is close to that of the centralized setting ($SC1$). It has some degradation compared to $SC1$ because some nodes may have dependency across non-adjacent CLNs which is not accounted for in the inter-cloudlet communication. However, the fully decentralized setting $SC2$ has more performance degradation as compared to $SC1$ and $SC3$. This is because the number of communication hops in $S2$ is restricted to the communication ranges of taxis in the wireless ad hoc network (This is a distance of 100 m).

\par Now, let us compare the performance of the proposed algorithm in the three settings to that of the baselines according to Table \ref{table1}. First, for the \textit{taxi-region} comparison case, the proposed algorithm has higher RMSE and MAE values compared to the other baselines. However, it has a lower MAPE than the baselines. This result is not conclusive since this comparison case compares different quantities in different settings (the proposed algorithm predicts on a taxi level, whereas the baselines predict on a region level), we only include this result for the reader's reference. The other two comparison cases are more reasonable in the sense that similar quantities are compared. In the region-region comparison case, the proposed algorithm significantly outperforms the baselines. For example, the proposed algorithm in the centralized setting achieves reductions of 45.4\%, 56.0\%, and 58.5\% in MSE, MAE, and MAPE compared to CCRNN \cite{ye2021coupled}. Similar reductions are obtained with the proposed algorithm in decentralized and semi-decentralized settings. A similar result is seen in the taxi-taxi comparison case, where the three metrics of the baselines become larger as they are operated at a taxi node level. As an example, the proposed algorithm in the centralized setting achieves reductions of 17.46\%, 28.79\%, and 50.31\% in MSE, MAE, and MAPE compared to CCRNN \cite{ye2021coupled}. A similar performance is obtained with the proposed algorithm in decentralized and semi-decentralized settings.

\begin{figure}[!htb]
\centering
\resizebox{0.99\columnwidth}{!}{
\begin{tabular}{c}
\includegraphics[width=12cm]{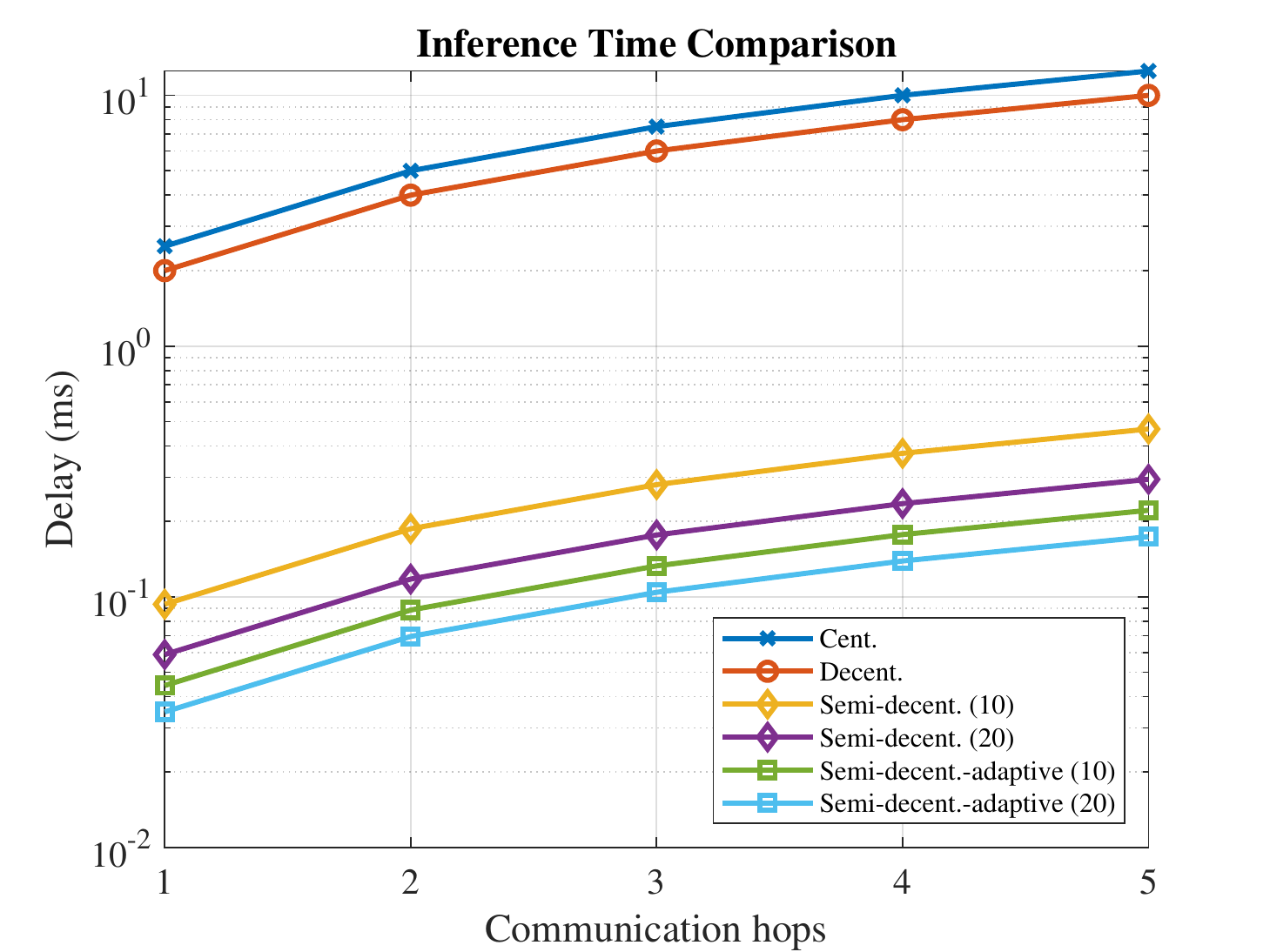}\\ 
\Large{(a)}\\
\includegraphics[width=12cm]{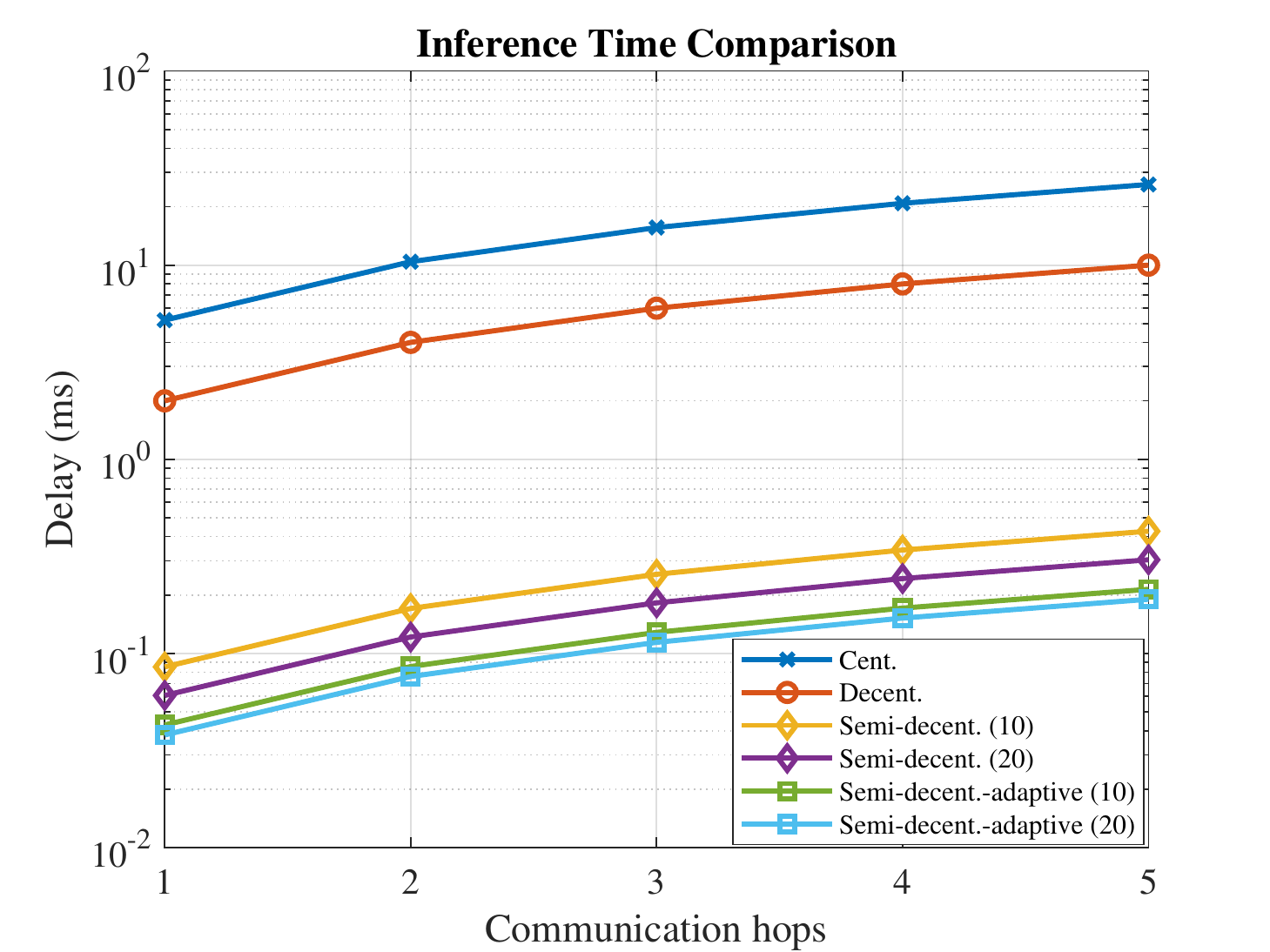}\\ \Large{(b)}
\end{tabular}}
\caption{Comparison of the overall inference time versus the number of message passing hops in four decentralization scenarios, with the proposed model in (a), and CCRNN in (b) .}
\label{inf_time_comp} 
\end{figure}

\subsection{The impact of decentralization on the GNN inference delay}
\par In this experiment, the objective is to compare the overall GNN inference time in centralized, decentralized, and semi-decentralized settings, calculated as detailed below.
\begin{itemize}[leftmargin=*]
\item In centralized GNN: The overall inference time equals the summation of; the time for nodes to upload their messages to the server, the time of inference for nodes using message passing by calculation and layer processing, and the time to send the eventual messages to their respective nodes. The communication medium between the nodes and the server assumed is the ITS-G5 standard from \cite{mannoni2019comparison}, and we adopt a packet transmission delay of 3.3 ms as reported in \cite{mannoni2019comparison}. As for the GNN layer computation time, we measure it as the execution time for one GNN layer. Similarly, measure the time required to pass the messages by computation instead of communication.

\item In decentralized GNN: The overall inference time for the nodes in the graph is calculated on a node level, where each node has its $L$-hop computational graph. We calculate the overall inference time for a given node by adding the transmission delays required to send and receive the messages to/from the nodes in its $L$-hop computational graph, and the processing time at each GNN layer. The communication medium assumed in this setting is an ad hoc wireless network, and we adopt empirical values for the transmission delay from \cite{miya2021experimental}. In this network, a source node forwards its message to relay nodes that forward it till it reaches the destination node. As described in \cite{miya2021experimental}, the processing delay for a source node is about 16.55 ms which accounts for queuing and processing delays, whereas the link transmission delay is 7 ms. However, the processing delay at a relay node is 11.65 ms. Thus, an empirical value for the transmission delay calculated in this manner is about (16.55+7) ms for source nodes and (11.65+7) ms for relay nodes. As for the layer processing time, a decentralized device is assumed to have a 100 times longer processing time as compared to a centralized cloud server, as commonly assumed in the literature \cite{challapalle2021crossbar}.

\item In semi-decentralized GNN: The overall inference time is calculated as the time for the slowest CLN to obtain the embeddings of its nodes. This is calculated as the sum of the time required to upload the node messages to the CLN, the time for the CLN to produce latent embeddings based on the initial node messages, and the time to send these latent embeddings to their nodes. Furthermore, if a CLN has boundary nodes connected to nodes in an adjacent CLN, then the two CLNs have to exchange the updated messages of their connected nodes for each GNN layer. Thus, the overall message passing delay per GNN layer is the sum of the message passing by calculation (relatively small) and the inter-CLN message passing (relatively large). As for the layer processing time, a cloudlet in the semi-decentralized setting is assumed to have an order of magnitude slower computation as compared to the cloud server in the centralized setting. It is noted that if an adaptive assignment is applied, then we add its time delay. This is the sum of time to send information from a CLN to its neighbor, the time for the receiving CLN to perform adaptive k-means assignment, and the time for sending the result to the other CLN. 
\end{itemize}

\par With the above specifications, we assume a total of 255 nodes spread in a city area. Then, we quantify the overall inference time over the number of communication hops for the following scenarios.
\begin{itemize}[leftmargin=*]
\item \textit{Cent.:} centralized inference as shown in Fig. \ref{threesettings}-a
\item \textit{Decent.:} decentralized inference as shown in Fig. \ref{threesettings}-b
\item \textit{Semi-decent.:} semi-decentralized inference, as shown in Fig. \ref{threesettings}-c, where decentralization is obtained using 10 CLNs of uniform node assignment. We also consider the case of 20 CLNs for observing the effect of the number of CLNs on performance. 

\item \textit{Semi-decent.-adaptive:} semi-decentralized inference where decentralization is obtained by splitting the nodes into the coverage of 10 non-uniform CLNs with the proposed adaptive assignment according to Algorithm \ref{Algorithm2}. Similar to the previous scenario, we also include the case of using 20 CLNs. 
\end{itemize}

\par Fig. \ref{inf_time_comp} shows the results of the above experiment, with the GNN model being the proposed hetGNN-LSTM model in (a), and CCRNN \cite{ye2021coupled} in (b). Several conclusions can be made from this figure. First, decentralization reduces the overall inference time (by comparing \textit{Cent.} and \textit{Decent.}). Second, the added benefit of semi-decentralization is shown in the significant reduction, of about an order of magnitude, in inference time attained by the proposed setting (\textit{Semi-decent.} and \textit{Semi-decent-adaptive.}) as compared to either centralization or decentralization. Furthermore, finer decentralization (20 CLNs) is shown more advantageous than coarse decentralization (10 CLNs). This is consistently seen in both \textit{Semi-decent.} and \textit{Semi-decent.-adaptive:}. Moreover, the advantage of adaptive node-CLN assignment is shown by a significant reduction of around 2-times in inference time in \textit{Semi-decent.-adaptive} compared to the case of uniform assignment \textit{Semi-decent}. This is consistent with the expected reduction of inter-CLN edge communication. 

\par
\begin{figure}[!htb]
\centering
\resizebox{0.8\columnwidth}{!}{
\begin{tabular}{c}
\includegraphics[width=12cm]{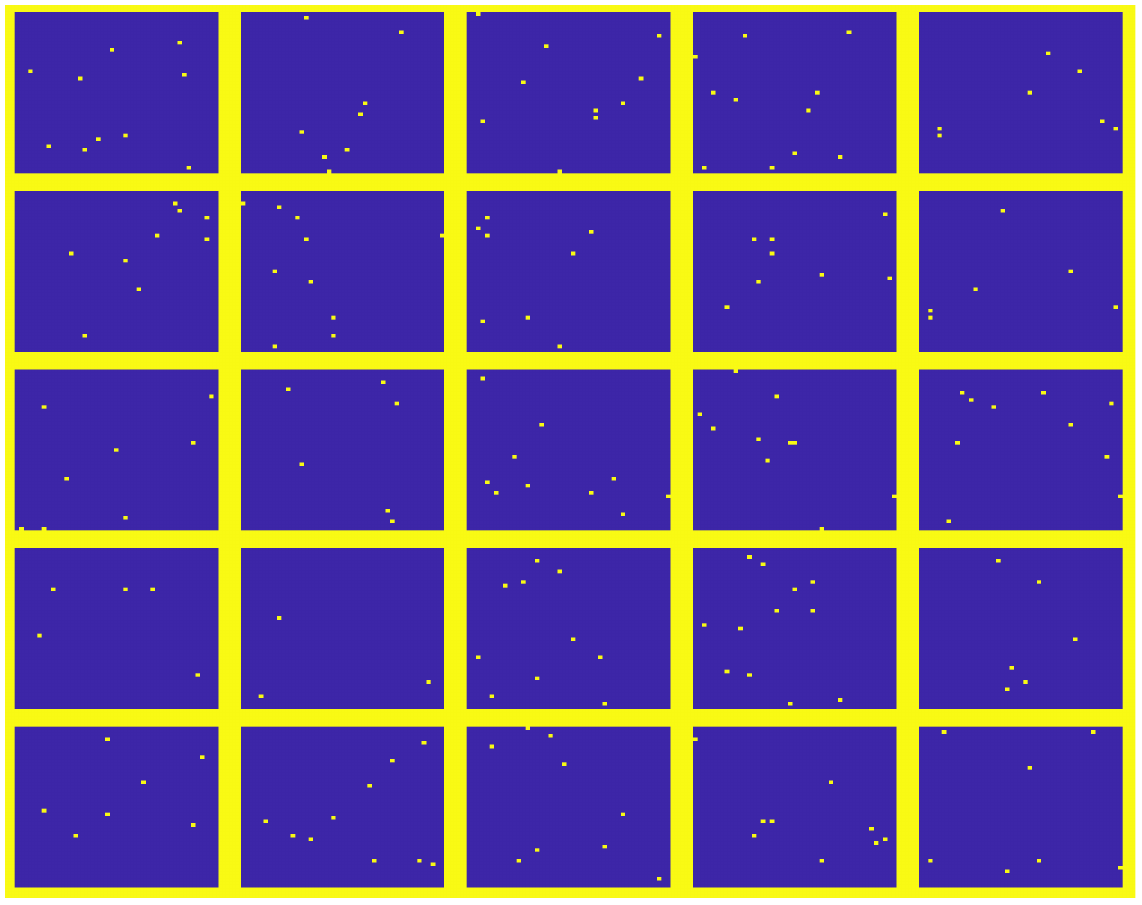}\\ 
\Large{(a)}\\
\includegraphics[width=12cm]{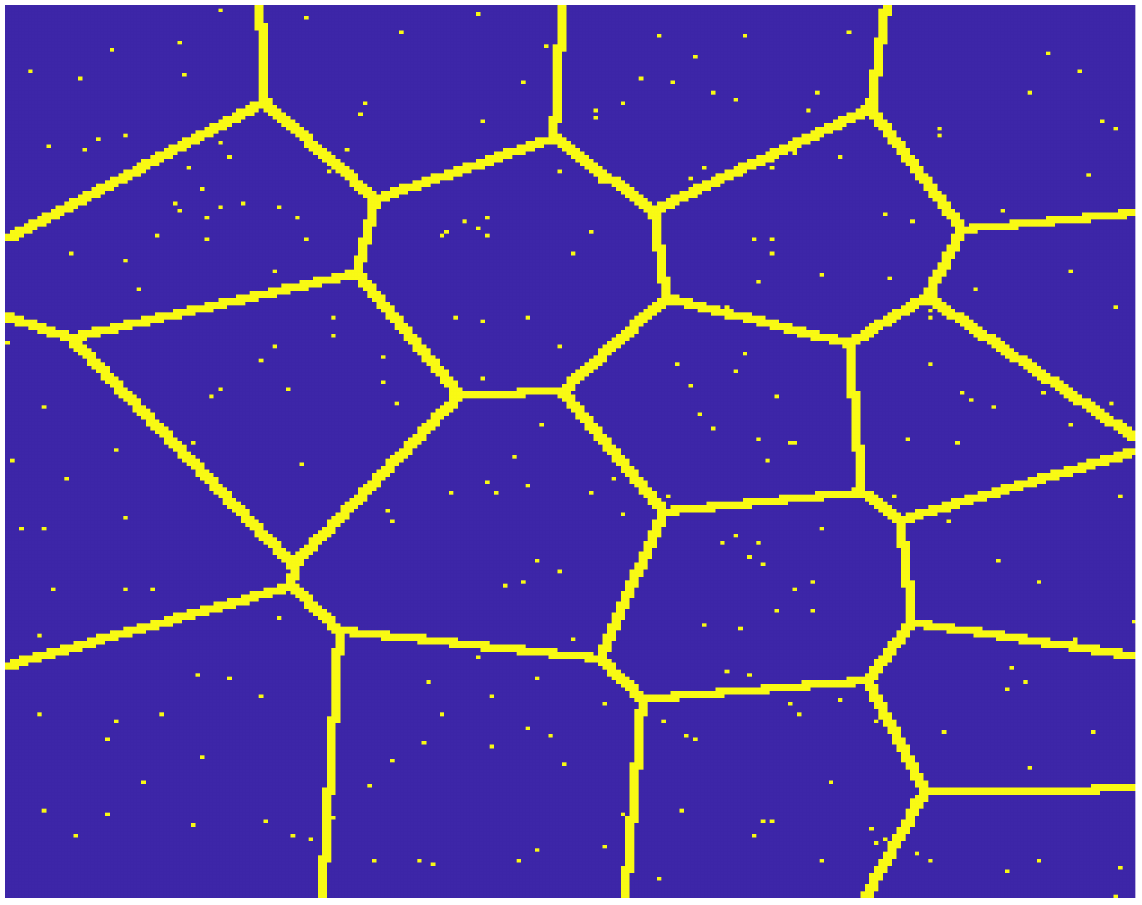}\\ \Large{(b)}
\end{tabular}}
\caption{Uniform and adaptive assignment samples in (a) and (b), respectively.}
\label{sample_assign} 
\end{figure}

 Fig. \ref{sample_assign} visually illustrates adaptive assignment by comparing a sample uniform assignment to an adaptive one in parts (a) and (b), respectively. In part (b), we obtain adaptive assignments using the proposed distributed adaptive assignment approach in Algorithm \ref{Algorithm2}. CLN boundaries are denoted by solid yellow lines and nodes are denoted by yellow dots. It can be seen that adaptive assignment can help reduce the number of taxis connected across CLNs, and thus, the volume of inter-CLN communication. This interprets the reduction in inference delay shown in the \textit{Semi-decent.-adaptive} scenario in Fig. \ref{inf_time_comp}.

\subsection{The generalizability of the hetGNN model}

\par
\begin{figure}[!htb]
\centering
\resizebox{0.99\columnwidth}{!}{
\includegraphics[width=12cm]{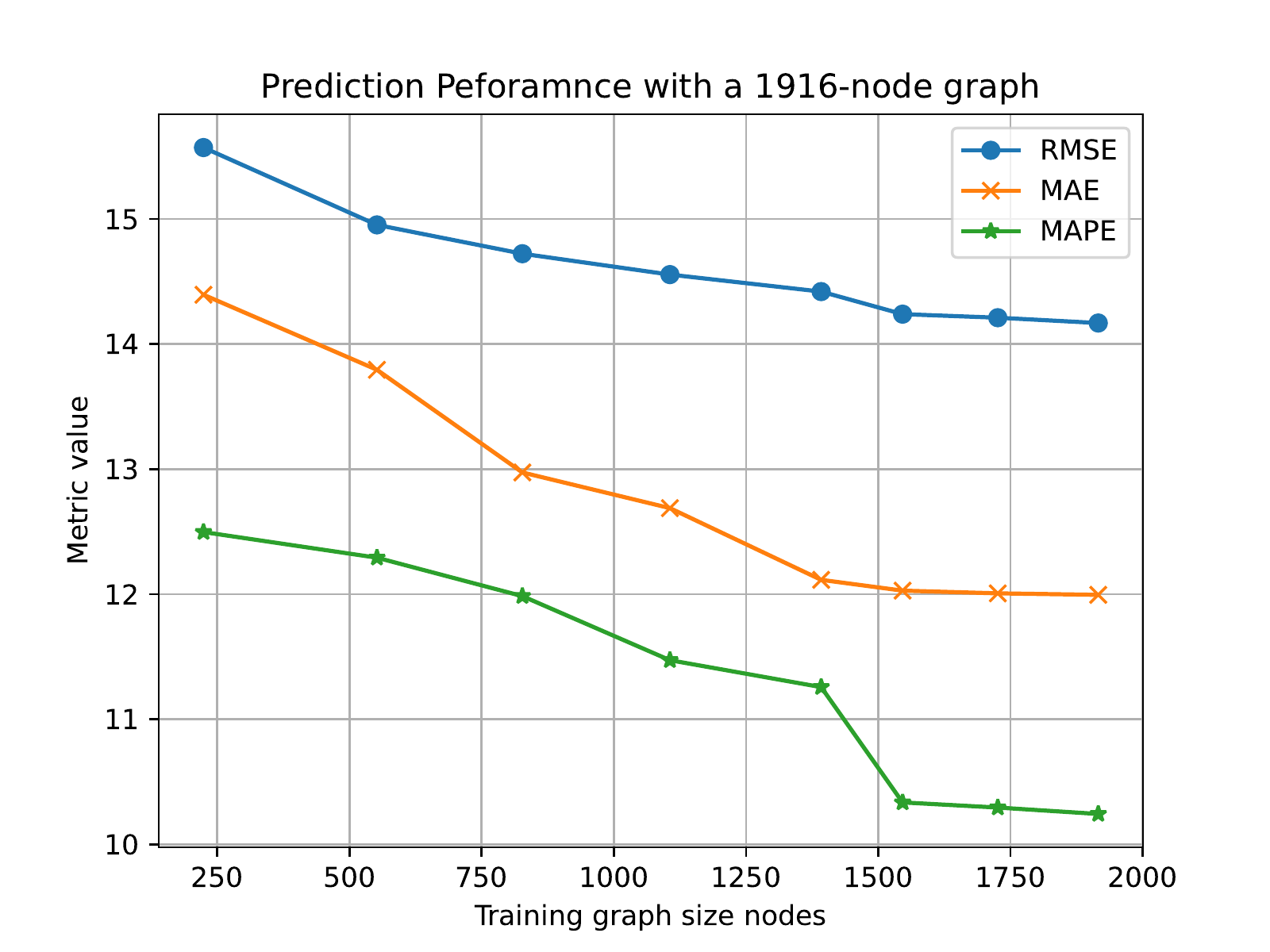}}
\caption{Quality metrics for testing with a fixed graph size while varying training graph sizes.}
\label{generalizability} 
\end{figure}

 A key advantage of GNNs is that even when trained with moderately sized graphs, they can generalize to larger graphs with unforeseen structures \cite{li2020graph}. In this experiment, we investigate the impact of training graph sizes on the performance of the proposed hetGNN-LSTM algorithm. For this purpose, we consider a sample test graph of 1916 nodes and test it with different models trained with gradually increasing the training graph sizes from 220 to 1916. Fig. \ref{generalizability} shows the testing RMSE, MAE, and MAPE performance metrics versus the training graph size.

\par In view of Fig. \ref{generalizability}, it can be seen that the drop in training set size does not incur a proportional drop in the performance as one moves from 1916 nodes to 220 nodes. With about one-tenth of the graph sizes for the training, the performance loss is still relatively marginal. This result establishes the generalizability of the hetGNN model.

\subsection{Training the model using federated learning}
\par
\begin{figure}[!htb]
\centering
\resizebox{0.99\columnwidth}{!}{
\includegraphics[width=12cm]{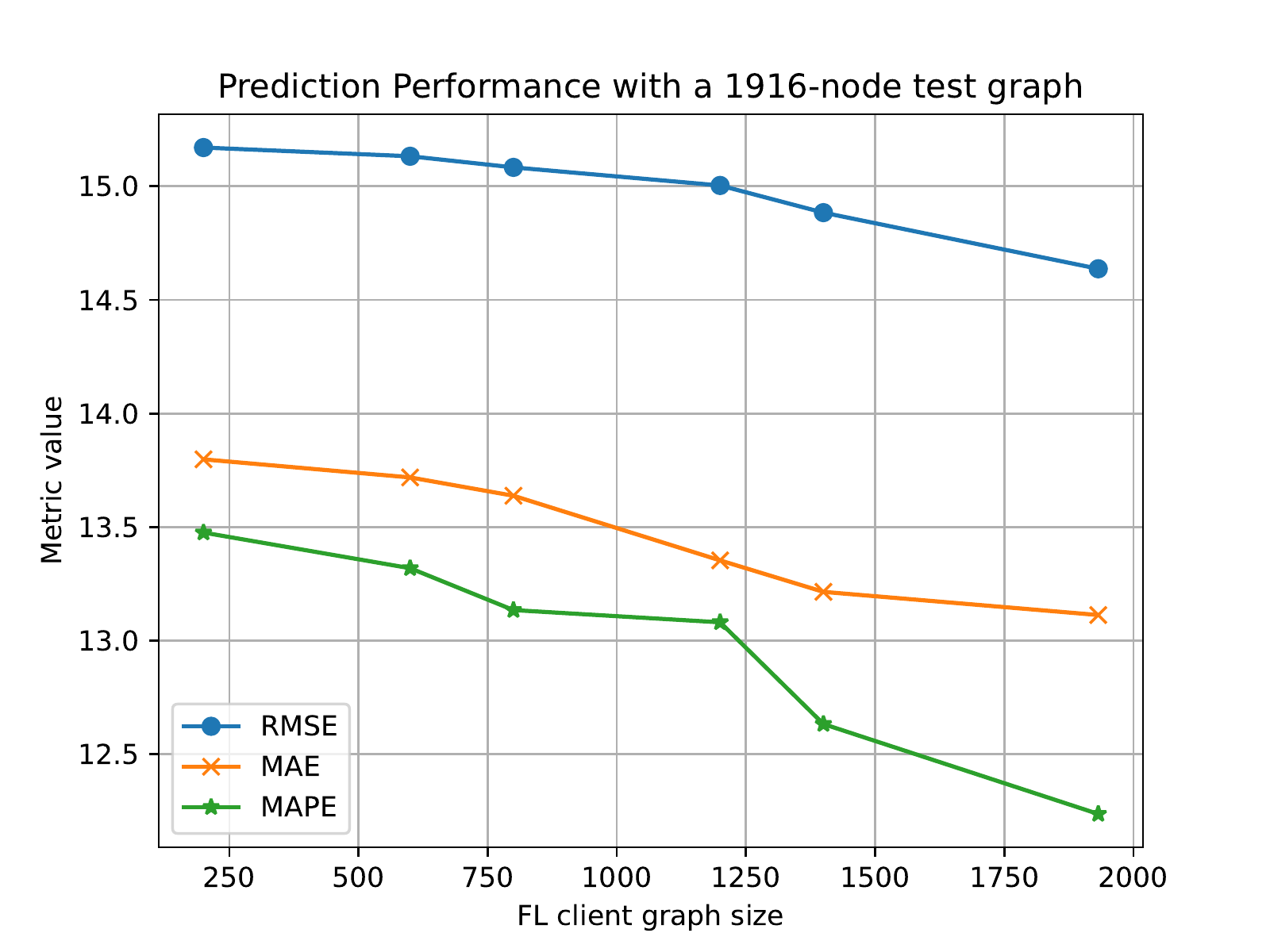}}
\caption{Quality metrics for testing with a fixed graph size and different average client graph sizes in distributed FL.}
\label{FL_testing} 
\end{figure}

 The above experiments focus on the inference of the proposed model. Still, it is interesting to examine its operation with distributed learning such as federated learning (FL) \cite{zhang2021federated,he2021fedgraphnn}. However, we assume a server-free FL framework due to the absence of a central server. This framework is similar to the server-free FL approach in \cite{he2019central} where neighboring clients exchange models, aggregate them to obtain a model initialization, train on local data, and then exchange the trained models, and so on. However, we assume that neighboring clients work with mutual trust. In this setting, each cloudlet is considered an FL client. Starting from the initial model states, each client exchanges its model with its adjacent neighbors, aggregates the received models along with its own local model, and trains on the aggregated model over its local data. Next, the trained models are again shared across direct neighbors and so on. This process is repeated for a predetermined number of FL rounds. In the following experiment, we divide the city area into 10 areas covered by 10 CLNs and apply this decentralized FL across them for 10 rounds. Table \ref{table3} lists several hyper-parameters and specifications for this experiment. We repeat this experiment with varying CLN graph sizes. For each case, we test the aggregate models over a test graph size of 1916 nodes and plot the average performance metrics across the 10 clients. Fig. \ref{FL_testing} shows the results of this experiment. 

\begin{table}[!ht]
\centering
\caption{Key federated learning parameters and specifications.}
\begin{tabular}{|l|l|}
\hline
Parameter/Property & Specification \\ \hline
No. of clients& 10 \\ \hline
No. or FL training rounds & 10 \\ \hline
Aggregation operator& FedAvg \\ \hline
Local training epochs & 50 \\ \hline
Local training learning rate & 0.005\\ \hline
\end{tabular}
\label{table3}
\end{table}
\par One can make the following observations in view of Fig. \ref{FL_testing}. First, an FL-trained model exhibits a slight degradation in its prediction quality as compared to a model trained in a centralized manner. This can be seen in view of the drop in the RMSE, MAE, and MAPE in Fig. \ref{FL_testing} compared to Fig. \ref{generalizability}, where the values in FL training are on average 7\% less than their corresponding values in centralized training. Second, it can be seen that the size of the client graph has a small impact on the quality of the aggregated model. This result is consistent with the generalizability of the GNN model established in the previous experiment. Therefore, one can still train with moderately sized graphs, possibly at the cloudlet area level without sacrificing the performance.

\begin{figure}[!thb]
\centering
\resizebox{0.95\columnwidth}{!}{
\includegraphics[width=12cm]{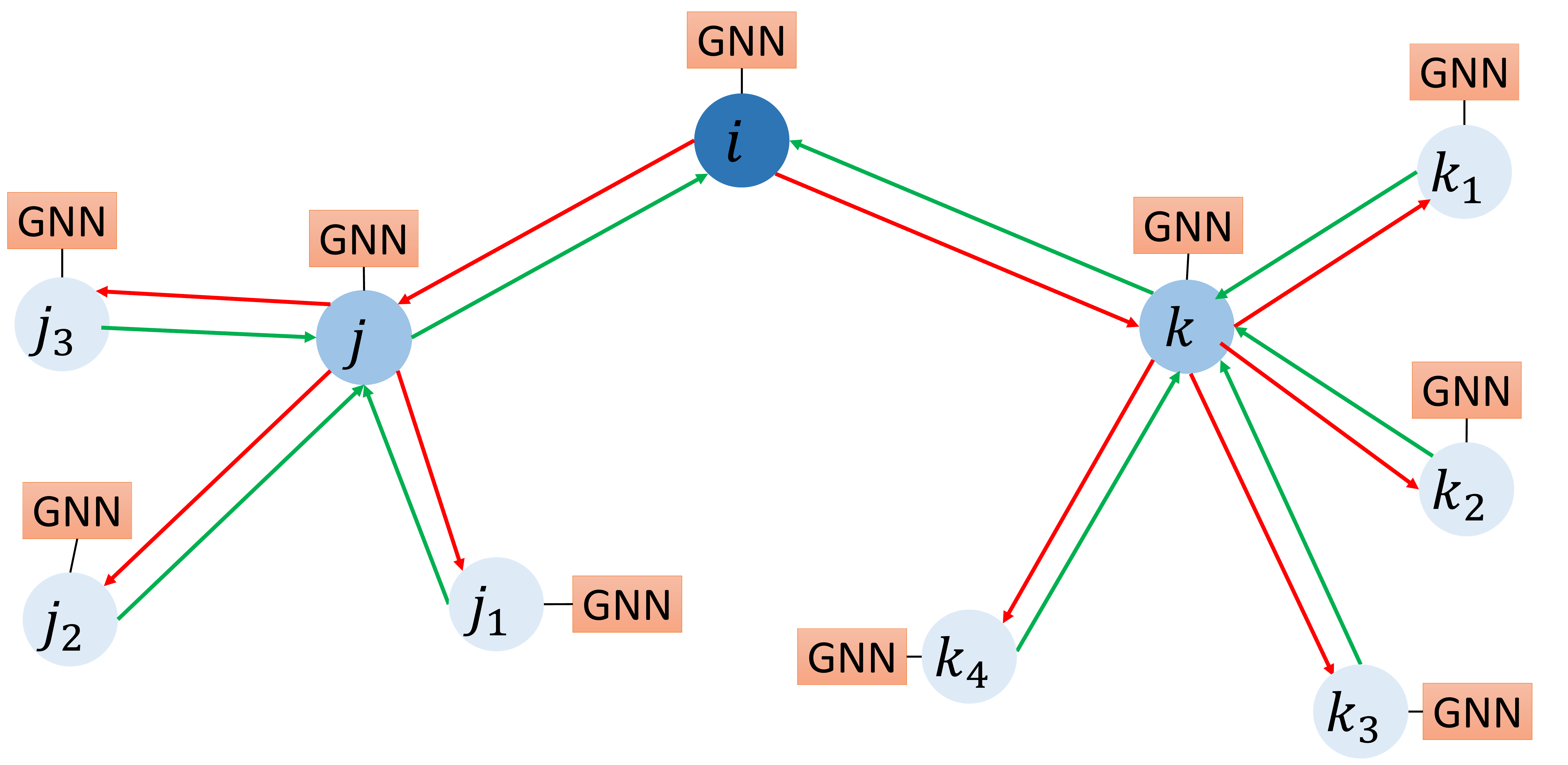}}
\caption{A sample node's computational graph in the decentralized GNN setting.}
\label{theorem_proof} 
\end{figure}

\section{Conclusion and Future Work} 
\label{Section5}
\par In this paper, we propose a hetGNN-LSTM algorithm for taxi demand and supply prediction making use of several edge types between graph nodes. To enable this approach for large-scale training and inference, we also propose a semi-decentralized GNN approach that resolves the scalability and excessive communication delay limitations of centralized and decentralized GNNs, respectively. We propose the use of multiple CLNs to enable this semi-decentralized approach. Through experiments using real data, the proposed hetGNN-LSTM algorithm is shown to be effective in predicting taxi demand and supply on a taxi level. This allows for dynamic graph evolution over time as opposed to the existing approaches assuming static graphs. The proposed hetGNN-LSTM model is shown to generalize well to larger graphs with unforeseen structures. Moreover, the proposed semi-decentralized approach allows for cloudlet-level federated learning without sacrificing performance.

\par Future extensions to this work will include incorporating other node and edge types on the constructed HIN. Besides, extensions will also include the development of custom-made hetGNN models to better fit the traffic demand prediction. Namely, by more explicit incorporation of time dependency. It is also interesting to devise better ways of node-CLN assignment. 

\section{Acknowledgment}
\par This work is supported in part by the National Science Foundation under Grant No. 2216772.

\appendix

\section{Appendix}
\label{Appendix}
\subsection{The proof of Theorem 1}
\begin{proof}

\par To derive the delay bounds, we quantify the inference delay bounds on a small graph with 2-hops and then generalize the derived bounds for $L$-hop graphs with arbitrary numbers of nodes. Consider the computational graph of node $i$ shown in Fig. \ref{theorem_proof}. This node has nodes $j$ and $k$ as its 1-hop neighbors, and nodes $j_1$ through $j_3$, and $k_1$ through $k_4$ as its hop-2 neighbors. Let us express the overall inference delay with messages from the 1-hop and 2-hop nodes to node $i$ according to the topology of this computational graph. We assume that nodes communicate through an ad hoc wireless network. Also, a node can only communicate with one node at a time. Let $t_s$ and $t_r$ denote the (per-node) sending and receiving transmission delays, respectively. For simplicity, let us further assume that nodes have similar separations for simplicity and $t_s=t_r$. Also, let $t_p$ denote the processing time for messages received at a node through a GNN layer.

\par The total 2-hop inference delay ($\Delta_{tot, 2}$) can be written as.
\begin{equation}
\Delta_{tot, 2}=\Delta_1 +\Delta_2
\end{equation}
\noindent where $\Delta_1$ is the time required to pass the 1-hop messages from $j$ and $k$ to $i$, process them by the first GNN layer, and send the processed message to $j$ and $k$. So, the total delay of hop-1 is: 
\begin{equation}
\Delta_1=2(t_s+tr)+tp=2 \times 2( t_r+t_p)
\end{equation}
\par In general, for node $i$ with degree $d_i$,
\begin{equation}
\Delta_1=2(t_s+tr)+tp=2 \times d_i( t_r+t_p)
\end{equation}
\noindent where $d_i$ is the degree of node $i$.

\par Let us consider the delay in the 2 communication hop. $\Delta_2$ is the time required to receive the messages from the neighbors of $j$ and $k$, sending them to $i$, processing them by the second GNN layer, and then sending the result back to the neighbors of $j$ and $k$. The time required for collecting the messages from the neighbors of $j$ (to itself) is $\Delta_2^j=d_j(t_s+t_r+t_p)=2d_j t_r+t_p$. Similarly, the time required to collect the messages from the neighbors of $k$ (to itself) is $\Delta_2^k=2d_kt_r+t_p$. These two messages need then to be sent to $i$. So, this requires another time delay of $\Delta_{j,k to i}=2d_i t_r$. 

\par To this end, the topology of the graph determines the way how the delays of $j$ and $k$ contribute to the overall delay. Specifically, if $j$ and $k$ have uncommon nodes in their 1-hop graph (except for $i$) then they can work concurrently on receiving their messages. Thus, their delay will be determined by the slowest among them. This is the minimum value of their delays. However, if they have common nodes, then they must sequentially communicate with these nodes. This elongates the delay. It can be maximally equal to the summation of delays $j$ and $k$. So, the delay for hop-2 message passing is.
\begin{equation}
\Delta_{tot, 2}=\Delta_1+\Delta_2^j+\Delta_2^k-\Delta_{conc}
\end{equation}
\noindent where $\Delta_{conc}$ is the time duration when $j$ and $k$ concurrently receive messages (from their exclusive neighbors). The minimum value of $\Delta_{conc}$ is zero when the $j$ and $k$ can not work simultaneously, i.e., when they share the same 1-hop neighborhood. This maximizes the delay to $\Delta_{2,max}=\Delta_1+\Delta_2^j+\Delta_2^k$. On the other hand, the minimum value of $\Delta_{conc}$ is $\min\{\Delta^j_2,\Delta^k_2\}$ when $j$ and $k$ can work completely simultaneously (except for communicating with node (i)). In that case, $\Delta_{2}$ is minimized to $\Delta_{2,min}=\Delta_1+\max\{\Delta^j_2,\Delta^k_2\}$.
\par Expanding $\Delta_2^j$ and $\Delta_2^k$ and adding common terms, and with the delay has the following bounds.
\begin{multline}
2t_r[2d_i+ \max_{j\in N_2(i)}\{d_x\}\}]+3t_p \\
\leq 
\Delta_{tot, 2} \leq \\
2t_r[2d_i+\sum_{x \in N_2(i)} d_x]+3t_p
\end{multline}
\par Following the same logic, we can generalize the above bounds for an $l$-hops since an $l$-hop message received at node $i$ is an $l-1$-hop message received at $j$ and $k$ then forwarded to $i$.
\begin{multline}
2t_r[ld_i+ \max_{x\in N_l(i)}\{d_x\} \}]+(l+1)t_p \\
\leq 
\Delta_{tot, l} \leq \\
2t_r[ld_i+\sum_{x \in N_l(i)} d_x]+(l+1)t_p
\end{multline}
\noindent where $N_l(i)$ denotes nodes connected to $i$ and $l$-hops always.

\par In an $L$-hop computational graph, the hop-1 messages will be exchanged $L$ times, the hop-2 messages will be exchanged $L-1$ times, and so on. So, the total $L$-hop delay ($\Delta_{tot, L}$ is within the following bounds.
\begin{multline}
2t_r\sum_{l=1}^L(L-l+1)[ld_i+\max_{ x\in N_l(i)} \{d_x\}+(l+1)t_p ]\\
\leq 
\Delta_{tot, L} 
\leq
\\
2t_r\sum_{l=1}^L(L-l+1)[ld_i+\sum_{x \in N_l(i)} d_x+(l+1)t_p]
\end{multline}
\end{proof}
\noindent These bounds have quadratic growth with the number of hops. 



\begin{thebibliography}{10}
\providecommand{\url}[1]{#1}
\csname url@samestyle\endcsname
\providecommand{\newblock}{\relax}
\providecommand{\bibinfo}[2]{#2}
\providecommand{\BIBentrySTDinterwordspacing}{\spaceskip=0pt\relax}
\providecommand{\BIBentryALTinterwordstretchfactor}{4}
\providecommand{\BIBentryALTinterwordspacing}{\spaceskip=\fontdimen2\font plus
\BIBentryALTinterwordstretchfactor\fontdimen3\font minus
  \fontdimen4\font\relax}
\providecommand{\BIBforeignlanguage}[2]{{%
\expandafter\ifx\csname l@#1\endcsname\relax
\typeout{** WARNING: IEEEtran.bst: No hyphenation pattern has been}%
\typeout{** loaded for the language `#1'. Using the pattern for}%
\typeout{** the default language instead.}%
\else
\language=\csname l@#1\endcsname
\fi
#2}}
\providecommand{\BIBdecl}{\relax}
\BIBdecl

\bibitem{li2012prediction}
X.~Li, G.~Pan, Z.~Wu, G.~Qi, S.~Li, D.~Zhang, W.~Zhang, and Z.~Wang,
  ``Prediction of urban human mobility using large-scale taxi traces and its
  applications,'' \emph{Frontiers of Computer Science}, vol.~6, no.~1, pp.
  111--121, 2012.

\bibitem{yu20193d}
B.~Yu, M.~Li, J.~Zhang, and Z.~Zhu, ``3d graph convolutional networks with
  temporal graphs: A spatial information free framework for traffic
  forecasting,'' \emph{arXiv preprint arXiv:1903.00919}, 2019.

\bibitem{moreira2013predicting}
L.~Moreira-Matias, J.~Gama, M.~Ferreira, J.~Mendes-Moreira, and L.~Damas,
  ``Predicting taxi--passenger demand using streaming data,'' \emph{IEEE
  Transactions on Intelligent Transportation Systems}, vol.~14, no.~3, pp.
  1393--1402, 2013.

\bibitem{taxi2017tlc}
N.~Taxi, L.~Commission \emph{et~al.}, ``Tlc trip record data,'' \emph{Accessed
  January 2023}, vol.~12, 2017.

\bibitem{chen2020multi}
W.~Chen, L.~Chen, Y.~Xie, W.~Cao, Y.~Gao, and X.~Feng, ``Multi-range attentive
  bicomponent graph convolutional network for traffic forecasting,'' in
  \emph{Proceedings of the AAAI conference on artificial intelligence},
  vol.~34, no.~04, 2020, pp. 3529--3536.

\bibitem{xu2019incorporating}
Y.~Xu and D.~Li, ``Incorporating graph attention and recurrent architectures
  for city-wide taxi demand prediction,'' \emph{ISPRS International Journal of
  Geo-Information}, vol.~8, no.~9, p. 414, 2019.

\bibitem{liu2019contextualized}
L.~Liu, Z.~Qiu, G.~Li, Q.~Wang, W.~Ouyang, and L.~Lin, ``Contextualized
  spatial--temporal network for taxi origin-destination demand prediction,''
  \emph{IEEE Transactions on Intelligent Transportation Systems}, vol.~20,
  no.~10, pp. 3875--3887, 2019.

\bibitem{davis2020grids}
N.~Davis, G.~Raina, and K.~Jagannathan, ``Grids versus graphs: Partitioning
  space for improved taxi demand-supply forecasts,'' \emph{IEEE Transactions on
  Intelligent Transportation Systems}, vol.~22, no.~10, pp. 6526--6535, 2020.

\bibitem{xu2022application}
J.-Y. Xu, S.~Zhang, C.-C. Wu, W.-C. Lin, and Q.-L. Yuan, ``Application of an
  adaptive adjacency matrix-based graph convolutional neural network in taxi
  demand forecasting,'' \emph{Mathematics}, vol.~10, no.~19, p. 3694, 2022.

\bibitem{tong2017simpler}
Y.~Tong, Y.~Chen, Z.~Zhou, L.~Chen, J.~Wang, Q.~Yang, J.~Ye, and W.~Lv, ``The
  simpler the better: a unified approach to predicting original taxi demands
  based on large-scale online platforms,'' in \emph{Proceedings of the 23rd ACM
  SIGKDD international conference on knowledge discovery and data mining},
  2017, pp. 1653--1662.

\bibitem{xu2017real}
J.~Xu, R.~Rahmatizadeh, L.~B{\"o}l{\"o}ni, and D.~Turgut, ``Real-time
  prediction of taxi demand using recurrent neural networks,'' \emph{IEEE
  Transactions on Intelligent Transportation Systems}, vol.~19, no.~8, pp.
  2572--2581, 2017.

\bibitem{wang2018deepstcl}
D.~Wang, Y.~Yang, and S.~Ning, ``Deepstcl: A deep spatio-temporal convlstm for
  travel demand prediction,'' in \emph{2018 international joint conference on
  neural networks (IJCNN)}.\hskip 1em plus 0.5em minus 0.4em\relax IEEE, 2018,
  pp. 1--8.

\bibitem{ke2017short}
J.~Ke, H.~Zheng, H.~Yang, and X.~M. Chen, ``Short-term forecasting of passenger
  demand under on-demand ride services: A spatio-temporal deep learning
  approach,'' \emph{Transportation research part C: Emerging technologies},
  vol.~85, pp. 591--608, 2017.

\bibitem{li2017diffusion}
Y.~Li, R.~Yu, C.~Shahabi, and Y.~Liu, ``Diffusion convolutional recurrent
  neural network: Data-driven traffic forecasting,'' \emph{arXiv preprint
  arXiv:1707.01926}, 2017.

\bibitem{wu2019graph}
Z.~Wu, S.~Pan, G.~Long, J.~Jiang, and C.~Zhang, ``Graph wavenet for deep
  spatial-temporal graph modeling,'' \emph{arXiv preprint arXiv:1906.00121},
  2019.

\bibitem{ye2021coupled}
J.~Ye, L.~Sun, B.~Du, Y.~Fu, and H.~Xiong, ``Coupled layer-wise graph
  convolution for transportation demand prediction,'' in \emph{Proceedings of
  the AAAI conference on artificial intelligence}, vol.~35, no.~5, 2021, pp.
  4617--4625.

\bibitem{hong2020heteta}
H.~Hong, Y.~Lin, X.~Yang, Z.~Li, K.~Fu, Z.~Wang, X.~Qie, and J.~Ye, ``Heteta:
  heterogeneous information network embedding for estimating time of arrival,''
  in \emph{Proceedings of the 26th ACM SIGKDD international conference on
  knowledge discovery \& data mining}, 2020, pp. 2444--2454.

\bibitem{lee2021decentralized}
M.~Lee, G.~Yu, and H.~Dai, ``Decentralized inference with graph neural networks
  in wireless communication systems,'' \emph{IEEE Transactions on Mobile
  Computing}, 2021.

\bibitem{ge2017energy}
X.~Ge, J.~Yang, H.~Gharavi, and Y.~Sun, ``Energy efficiency challenges of 5g
  small cell networks,'' \emph{IEEE communications Magazine}, vol.~55, no.~5,
  pp. 184--191, 2017.

\bibitem{hochreiter1997long}
S.~Hochreiter and J.~Schmidhuber, ``Long short-term memory,'' \emph{Neural
  computation}, vol.~9, no.~8, pp. 1735--1780, 1997.

\bibitem{niu2019predicting}
K.~Niu, C.~Wang, X.~Zhou, and T.~Zhou, ``Predicting ride-hailing service demand
  via rpa-lstm,'' \emph{IEEE Transactions on Vehicular Technology}, vol.~68,
  no.~5, pp. 4213--4222, 2019.

\bibitem{niu2018real}
K.~Niu, C.~Cheng, J.~Chang, H.~Zhang, and T.~Zhou, ``Real-time taxi-passenger
  prediction with l-cnn,'' \emph{IEEE Transactions on Vehicular Technology},
  vol.~68, no.~5, pp. 4122--4129, 2018.

\bibitem{feng2021using}
Y.~Feng, T.~Zhang, A.~P. Sah, L.~Han, and Z.~Zhang, ``Using appearance to
  predict pedestrian trajectories through disparity-guided attention and
  convolutional lstm,'' \emph{IEEE Transactions on Vehicular Technology},
  vol.~70, no.~8, pp. 7480--7494, 2021.

\bibitem{meng2023trajectory}
Q.~Meng, H.~Guo, Y.~Liu, H.~Chen, and D.~Cao, ``Trajectory prediction for
  automated vehicles on roads with lanes partially covered by ice or snow,''
  \emph{IEEE Transactions on Vehicular Technology}, 2023.

\bibitem{hamilton2017representation}
W.~L. Hamilton, R.~Ying, and J.~Leskovec, ``Representation learning on graphs:
  Methods and applications,'' \emph{arXiv preprint arXiv:1709.05584}, 2017.

\bibitem{sahu2017ubiquity}
S.~Sahu \emph{et~al.}, ``The ubiquity of large graphs and surprising challenges
  of graph processing,'' \emph{Proceedings of the VLDB Endowment}, vol.~11,
  no.~4, pp. 420--431, 2017.

\bibitem{kiningham2022grip}
K.~Kiningham, P.~Levis, and C.~R{\'e}, ``Grip: A graph neural network
  accelerator architecture,'' \emph{IEEE Transactions on Computers}, 2022.

\bibitem{yan2020hygcn}
M.~Yan \emph{et~al.}, ``Hygcn: A gcn accelerator with hybrid architecture,'' in
  \emph{HPCA}.\hskip 1em plus 0.5em minus 0.4em\relax IEEE, 2020, pp. 15--29.

\bibitem{liang2023semantics}
G.~Liang, U.~Kintak, X.~Ning, P.~Tiwari, S.~Nowaczyk, and N.~Kumar,
  ``Semantics-aware dynamic graph convolutional network for traffic flow
  forecasting,'' \emph{IEEE Transactions on Vehicular Technology}, 2023.

\bibitem{zhang2019heterogeneous}
C.~Zhang, D.~Song, C.~Huang, A.~Swami, and N.~V. Chawla, ``Heterogeneous graph
  neural network,'' in \emph{Proceedings of the 25th ACM SIGKDD international
  conference on knowledge discovery \& data mining}, 2019, pp. 793--803.

\bibitem{hu2020heterogeneous}
Z.~Hu, Y.~Dong, K.~Wang, and Y.~Sun, ``Heterogeneous graph transformer,'' in
  \emph{Proceedings of The Web Conference 2020}, 2020, pp. 2704--2710.

\bibitem{sun2011pathsim}
Y.~Sun, J.~Han, X.~Yan, P.~S. Yu, and T.~Wu, ``Pathsim: Meta path-based top-k
  similarity search in heterogeneous information networks,'' \emph{Proceedings
  of the VLDB Endowment}, vol.~4, no.~11, pp. 992--1003, 2011.

\bibitem{guan2022personalized}
W.~Guan, F.~Jiao, X.~Song, H.~Wen, C.-H. Yeh, and X.~Chang, ``Personalized
  fashion compatibility modeling via metapath-guided heterogeneous graph
  learning,'' in \emph{Proceedings of the 45th International ACM SIGIR
  Conference on Research and Development in Information Retrieval}, 2022, pp.
  482--491.

\bibitem{wang2021self}
X.~Wang, N.~Liu, H.~Han, and C.~Shi, ``Self-supervised heterogeneous graph
  neural network with co-contrastive learning,'' in \emph{Proceedings of the
  27th ACM SIGKDD conference on knowledge discovery \& data mining}, 2021, pp.
  1726--1736.

\bibitem{zheng2021multi}
J.~Zheng, Q.~Ma, H.~Gu, and Z.~Zheng, ``Multi-view denoising graph
  auto-encoders on heterogeneous information networks for cold-start
  recommendation,'' in \emph{Proceedings of the 27th ACM SIGKDD Conference on
  Knowledge Discovery \& Data Mining}, 2021, pp. 2338--2348.

\bibitem{liu2018heterogeneous}
Z.~Liu, C.~Chen, X.~Yang, J.~Zhou, X.~Li, and L.~Song, ``Heterogeneous graph
  neural networks for malicious account detection,'' in \emph{Proceedings of
  the 27th ACM International Conference on Information and Knowledge
  Management}, 2018, pp. 2077--2085.

\bibitem{qian2021distilling}
Y.~Qian, Y.~Zhang, Y.~Ye, C.~Zhang \emph{et~al.}, ``Distilling meta knowledge
  on heterogeneous graph for illicit drug trafficker detection on social
  media,'' \emph{Advances in Neural Information Processing Systems}, vol.~34,
  pp. 26\,911--26\,923, 2021.

\bibitem{pujol2022unveiling}
D.~Pujol-Perich, J.~Su{\'a}rez-Varela, A.~Cabellos-Aparicio, and P.~Barlet-Ros,
  ``Unveiling the potential of graph neural networks for robust intrusion
  detection,'' \emph{ACM SIGMETRICS Performance Evaluation Review}, vol.~49,
  no.~4, 2022.

\bibitem{hamilton2017inductive}
W.~Hamilton, Z.~Ying, and J.~Leskovec, ``Inductive representation learning on
  large graphs,'' \emph{Advances in neural information processing systems},
  vol.~30, 2017.

\bibitem{zheng2020distdgl}
D.~Zheng \emph{et~al.}, ``Distdgl: distributed graph neural network training
  for billion-scale graphs,'' in \emph{IA3}.\hskip 1em plus 0.5em minus
  0.4em\relax IEEE, 2020, pp. 36--44.

\bibitem{zeng2022gnn}
L.~Zeng, C.~Yang, P.~Huang, Z.~Zhou, S.~Yu, and X.~Chen, ``Gnn at the edge:
  Cost-efficient graph neural network processing over distributed edge
  servers,'' \emph{IEEE Journal on Selected Areas in Communications}, 2022.

\bibitem{li2020graph}
Q.~Li \emph{et~al.}, ``Graph neural networks for decentralized multi-robot path
  planning,'' in \emph{IROS}.\hskip 1em plus 0.5em minus 0.4em\relax IEEE,
  2020, pp. 11\,785--11\,792.

\bibitem{tolstaya2020learning}
E.~Tolstaya \emph{et~al.}, ``Learning decentralized controllers for robot
  swarms with graph neural networks,'' in \emph{Conference on robot
  learning}.\hskip 1em plus 0.5em minus 0.4em\relax PMLR, 2020, pp. 671--682.

\bibitem{mannoni2019comparison}
V.~Mannoni, V.~Berg, S.~Sesia, and E.~Perraud, ``A comparison of the v2x
  communication systems: Its-g5 and c-v2x,'' in \emph{2019 IEEE 89th Vehicular
  Technology Conference (VTC2019-Spring)}.\hskip 1em plus 0.5em minus
  0.4em\relax IEEE, 2019, pp. 1--5.

\bibitem{miya2021experimental}
T.~Miya, K.~Ohshima, Y.~Kitaguchi, and K.~Yamaoka, ``Experimental analysis of
  communication relaying delay in low-energy ad-hoc networks,'' in \emph{2021
  IEEE 18th Annual Consumer Communications \& Networking Conference
  (CCNC)}.\hskip 1em plus 0.5em minus 0.4em\relax IEEE, 2021, pp. 1--2.

\bibitem{ao2012bounds}
W.~C. Ao and K.-C. Chen, ``Bounds and exact mean node degree and node isolation
  probability in interference-limited wireless ad hoc networks with general
  fading,'' \emph{IEEE transactions on vehicular technology}, vol.~61, no.~5,
  pp. 2342--2348, 2012.

\bibitem{karypis1998fast}
G.~Karypis and V.~Kumar, ``A fast and high quality multilevel scheme for
  partitioning irregular graphs,'' \emph{SIAM Journal on scientific Computing},
  vol.~20, no.~1, pp. 359--392, 1998.

\bibitem{karypis1997metis}
------, \emph{METIS: A software package for partitioning unstructured graphs,
  partitioning meshes, and computing fill-reducing orderings of sparse
  matrices}, 1997.

\bibitem{fey2019fast}
M.~Fey and J.~E. Lenssen, ``Fast graph representation learning with pytorch
  geometric,'' \emph{arXiv preprint arXiv:1903.02428}, 2019.

\bibitem{challapalle2021crossbar}
N.~Challapalle, K.~Swaminathan, N.~Chandramoorthy, and V.~Narayanan, ``Crossbar
  based processing in memory accelerator architecture for graph convolutional
  networks,'' in \emph{2021 IEEE/ACM International Conference On Computer Aided
  Design (ICCAD)}.\hskip 1em plus 0.5em minus 0.4em\relax IEEE, 2021, pp. 1--9.

\bibitem{zhang2021federated}
H.~Zhang, T.~Shen, F.~Wu, M.~Yin, H.~Yang, and C.~Wu, ``Federated graph
  learning--a position paper,'' \emph{arXiv preprint arXiv:2105.11099}, 2021.

\bibitem{he2021fedgraphnn}
C.~He, K.~Balasubramanian, E.~Ceyani, C.~Yang, H.~Xie, L.~Sun, L.~He, L.~Yang,
  P.~S. Yu, Y.~Rong \emph{et~al.}, ``Fedgraphnn: A federated learning system
  and benchmark for graph neural networks,'' \emph{arXiv preprint
  arXiv:2104.07145}, 2021.

\bibitem{he2019central}
C.~He, C.~Tan, H.~Tang, S.~Qiu, and J.~Liu, ``Central server free federated
  learning over single-sided trust social networks,'' \emph{arXiv preprint
  arXiv:1910.04956}, 2019.

\end{thebibliography}
\end{document}